\documentclass{article}

\usepackage[numbers]{natbib}
\usepackage[final]{neurips_2025}

\usepackage[utf8]{inputenc} % allow utf-8 input
\usepackage[T1]{fontenc}    % use 8-bit T1 fonts
\usepackage{hyperref}       % hyperlinks
\usepackage{url}            % simple URL typesetting
\usepackage{booktabs}       % professional-quality tables
\usepackage{amsfonts}       % blackboard math symbols
\usepackage{nicefrac}       % compact symbols for 1/2, etc.
\usepackage{microtype}      % microtypography
\usepackage{xcolor}         % colors
\usepackage{graphicx}
\usepackage{multirow}
\usepackage{amsmath,amssymb,amsfonts}
\usepackage{mathrsfs}
\usepackage{textcomp}
\usepackage{manyfoot}
\usepackage{booktabs}
\usepackage{algorithm}
\usepackage{algorithmicx}
\usepackage{algpseudocode}
\usepackage{listings}
\usepackage[skins]{tcolorbox}
\usepackage{fbox}
\usepackage{makecell}
\usepackage{multirow}
\usepackage{inconsolata}
\usepackage{colortbl}
\usepackage{threeparttable}
\usepackage{footnote}
\usepackage{tablefootnote}
\usepackage{float}
\usepackage{color}
\usepackage[misc]{ifsym}
\usepackage{soul}
\usepackage{titlesec}
\usepackage{wrapfig}
\usepackage{subcaption}
\usepackage{arydshln}
\usepackage{CJKutf8}
\usepackage{CJKspace}
\usepackage{CJKpunct}
\usepackage{booktabs}
\usepackage{pifont}
\graphicspath{{figures/}}

\makeatletter
\renewcommand{\@notice}{}
\makeatother

\usepackage{etoolbox}
\makeatletter
\AfterEndEnvironment{algorithm}{\let\@algcomment\relax}
\AtEndEnvironment{algorithm}{\kern2pt\hrule\relax\vskip3pt\@algcomment}
\let\@algcomment\relax
\newcommand\algcomment[1]{\def\@algcomment{\footnotesize#1}}
\renewcommand\fs@ruled{\def\@fs@cfont{\bfseries}\let\@fs@capt\floatc@ruled
	\def\@fs@pre{\hrule height.8pt depth0pt \kern2pt}%
	\def\@fs@post{}%
	\def\@fs@mid{\kern2pt\hrule\kern2pt}%
	\let\@fs@iftopcapt\iftrue}
\makeatother

\makeatletter
\def\whline#1{%
	\noalign{\ifnum0=`}\fi\hrule \@height #1 \futurelet
	\reserved@a\@xhline}

\usepackage{paralist}

\definecolor{gray1}{rgb}{0.93,0.93,0.93}
\definecolor{darkgreen}{rgb}{0.21,0.52,0.11}
\definecolor{skyblue}{RGB}{135,206,235}
\definecolor{dodgerblue}{RGB}{30,144,255}
\definecolor{lightblue}{RGB}{229,248,255}
\definecolor{cvprblue}{rgb}{0.21,0.49,0.74}
\definecolor{blue1}{RGB}{0,112,192}
\definecolor{deepred}{RGB}{170,54,2}
\definecolor{purple}{RGB}{0,86,84}

\title{OCRGenBench: A Comprehensive Benchmark for Evaluating OCR Generative Capabilities}

\author{
	Peirong Zhang$^1$,
	Haowei Xu$^1$,
	Jiaxin Zhang$^1$,
	Xuhan Zheng$^1$,
	Guitao Xu$^1$,\\
	\textbf{Yuyi Zhang}$^1$,
	\textbf{Junle Liu}$^1$,
	\textbf{Zhenhua Yang}$^1$,
	\textbf{Wei Zhou}$^2$,
	\textbf{Lianwen Jin}$^{1*}$ \\
	$^1$ South China University of Technology, Guangzhou 510641, China \\
	$^2$ Cardiff University, Cardiff CF24 4AG, UK \\
	\texttt{eeprzhang@mail.scut.edu.cn},
	\texttt{eelwjin@scut.edu.cn}
}

\begin{document}

\maketitle

\begin{abstract}
Improving visual text synthesis has long been a challenging and evolving frontier for image generation models. While recent state-of-the-art (SOTA) models have made remarkable strides in text generation capabilities, existing benchmarks inadequately assess their true performance due to narrow scope (scene text and posters only), isolated evaluation (T2I generation or editing separately), and insufficient difficulty (lacking challenging scenarios). To bridge this gap, we pioneer the unification of text-centric T2I generation, text editing, and OCR-related image-to-image translation to evaluate a model's holistic visual text synthesis abilities, \emph{i.e.}, \textbf{OCR generative capabilities}. Accordingly, we propose \textbf{OCRGenBench}, the most comprehensive benchmark to date for evaluating these abilities. OCRGenBench covers five common text categories and 33 OCR generative tasks, encompassing T2I generation, text editing, and other image-to-image OCR tasks (\emph{e.g.}, document dewarping and handwriting removal). The benchmark includes 1,060 human-annotated samples consisting of instruction-image-GT triplets, deliberately featuring high text density, diverse generation scales, varied aspect ratios, and bilingual content to capture real-world complexity. To enable holistic evaluation, we further propose OCRGenScore, a unified metric that comprehensively integrates evaluation of text accuracy, aesthetic quality, and instruction following. Extensive experiments on 19 cutting-edge closed-source and open-source generative models reveal that most models score below 60 out of 100. Through in-depth analysis, we identify critical limitations of current models, such as poor text localization, unintended content modifications, and inadequate handling of dense or small-scale text, which existing benchmarks have overlooked. We hope OCRGenBench establishes a robust standard for the evaluation of OCR generative capabilities, driving the evolution of generative models toward robust, reliable visual text synthesis in real-world scenarios. The benchmark and evaluation code are available at \url{https://github.com/NiceRingNode/Awesome-Generative-Models-for-OCR}.
\end{abstract}

\section{Introduction}
Generating images with machines represents humanity's ambitious pursuit to teach computers the fundamental human skill of creation, translating the visual world into algorithms. This field has seen remarkable progress from static ASCII art in the 1960s \cite{bell2016start,goodchild2021,computerhistory2024} to modern data-driven AI approaches, such as Variational AutoEncoders (VAE) \cite{vae2013kingma} and Generative Adversarial Networks (GAN) \cite{gan2014ian}. Recently, fueled by large-scale datasets and advanced architectures (\emph{e.g.}, large language models, diffusion models), instruction-based image generation and editing \cite{t2isurvey2024,editingsurvey2025huang} have proliferated. This paradigm enables users, even those non-technical, to generate visual content with natural language, achieving unprecedented flexible and controllable image synthesis.

Text images, such as documents, scene texts, and handwriting, serve as critical information carriers of human communication. Unlike natural objects, text images contain structured linguistic information that requires both visual perception and language understanding for proper interpretation. This dual requirement positions text image generation and editing as a particularly challenging frontier in generative AI \cite{elanwar2025generative,fontstyletransfer2025tai}. Despite significant advances in natural scene generation, generating high-quality text images remains problematic due to fundamental challenges, such as ensuring precise character formation and legibility across scales, maintaining semantic consistency between text and visual content, and handling diverse layouts.

Early general-domain generators excel at natural image generation but usually struggle with text image generation and editing \cite{sdxl2023podell,dalle32023,midjourney}, with few formally addressing this deficiency. This limitation spurs specialized text image models \cite{anytext2024tuo,condhand2023zhu,slogan2022luo,dreamtext2025wang}, which enhance text image generation at the cost of general generation abilities. Recently, recognizing text synthesis as an unresolved yet critical issue, researchers have sought to internalize high-quality text synthesis as a foundational skill of generative models, rather than delegating it to specialized solutions. Many state-of-the-art (SOTA) general-domain models, such as Nano-Banana-Pro \cite{nanobananapro2025}, GPT Image 1.5 \cite{gptimage152025}, and Qwen-Image \cite{qwenimage2025wu} have demonstrated strong text generation and editing capabilities without sacrificing general synthesis quality.

\begin{figure*}[t]
	\centering
	\includegraphics[width=0.93\textwidth]{mindmap.png}
	\caption{Task categorization and subordination of OCRGenBench. It includes 33 OCR generative tasks under five text categories: \textbf{document}, \textbf{handwriting}, \textbf{scene text}, \textbf{artistic text}, and \textbf{layout-rich text}. Each primary category encompasses multiple sub-tasks.}
	\label{Fig::mindmap}
\end{figure*}

To evaluate generative models' text-to-image (T2I) generation and image editing abilities, numerous general-domain benchmarks \cite{geneval2023,imgedit2025ye,longtext2025geng,oneigbench2025} have been proposed. However, text-related samples in these datasets are scarce (\emph{e.g.}, only 21 text prompts out of 200 in DrawBench \cite{imagen2022}, and 12 text editing prompts out of 734 in ImgEdit-Bench \cite{imgedit2025ye}). Some benchmarks \cite{anytext2024tuo,cvtg2k2025,texteditbench2025gui,vtpbench2025} tailored to visual text synthesis have emerged, such as AnyText-Benchmark \cite{anytext2024tuo} for multi-language text generation and TextEditBench \cite{texteditbench2025gui} for reasoning-intensive text editing. However, these benchmarks suffer from several limitations (1) \textbf{Limited text scenarios}. Most of them are confined to scene text and poster, neglecting model performance on other commonly encountered text images, such as document and handwriting. This can create blind spots where models excelling in one domain may fail catastrophically in others. (2) \textbf{Restricted task coverage}. Current benchmarks predominantly focus on T2I generation or editing, overlooking existing optical character recognition (OCR) image-to-image (I2I) translation tasks that require generating or reconstructing textual content with fine-grained text localization, style preservation, and visual cohesion (\emph{e.g.}, document deshadowing \cite{docres2024zhang} and scene text removal \cite{erasenet2020liu}). This narrow focus provides an incomplete picture of models' holistic text synthesis abilities. (3) \textbf{Insufficient difficulty}. Existing benchmarks typically feature minimal (word- or line-level) text content in generation and manipulate images with limited text during editing, failing to stress-test models' performance boundaries under complex and challenging conditions. These limitations collectively expose a critical \textbf{gap}: no existing benchmark provides a holistic, standardized assessment across diverse text scenarios, task types, and difficulty levels for visual text synthesis. Given the recent surge in models claiming strong text generation capabilities \cite{qwenimage2025wu,nanobananapro2025,zimage2025,ovis2025wang}, a more robust evaluation benchmark is urgently needed.

To this end, we propose \textbf{OCRGenBench}, a comprehensive benchmark for assessing current generators' visual text synthesis skills. OCRGenBench pioneers the unification of text-centric T2I generation, text editing, and OCR I2I translation tasks to reflect a model's overall visual text synthesis abilities, \emph{i.e.}, \textbf{OCR generative capabilities}. The benchmark spans five common textual content categories, including \emph{document}, \emph{handwriting}, \emph{scene text}, \emph{artistic text}, and \emph{layout-rich text}, evaluated through 33 representative OCR generative tasks, including \emph{T2I generation}, \emph{text editing}, \emph{document dewarping}, \emph{handwriting removal}, \emph{etc.} A detailed categorization is illustrated in Fig.~\ref{Fig::mindmap}. It contains 1,060 high-quality, manually annotated samples, encompassing ground truth (GT)-free generation prompts for T2I generation and instruction-image-GT triplets for image editing and OCR I2I tasks. Notably, we meticulously include diverse and challenging scenarios, such as page-level text generation, editing documents with dense text, and complicated font synthesis. Both English and Chinese prompts and images are curated to enable bilingual evaluation. In addition, due to the lack of metrics that can straightforwardly indicate unified OCR generative performance, we design a new metric termed \textbf{OCRGenScore}. Specifically, OCRGenScore is computed across the task dimension. For T2I generation and text editing tasks, we evaluate text content accuracy, structural consistency, and instruction following; for tasks with established metrics \cite{msssim2003,dd2025zhang}, we directly borrow standard practices. All task-specific metrics are then normalized to the 0-1 range and averaged to yield the final OCRGenScore. By consolidating OCRGenBench and OCRGenScore, we construct a fully automatic evaluation framework for OCR generative tasks, providing a comprehensive and robust assessment of generation models' OCR generative capabilities.

In experiments, we benchmark 19 SOTA generative models, including both specialized generators and unified understanding and generation models from closed-source and open-source domains. Results demonstrate that only two models, \emph{i.e.}, Nano Banana Pro and Flux.2-dev, achieve OCRGenScore over 70 (77.19 and 70.19, respectively), with most methods scoring below 60. Through in-depth analysis across different tasks and text categories, we identify several prevalent and critical issues of current models, such as poor text localization, unintended changes of surrounding text, and insufficient encoding and decoding granularity of small text. Language-specific comparisons reveal that models perform generally better on English than on Chinese text. These observations demonstrate that current models face significant challenges not only in typical text image generation and editing, but also in broader OCR generative tasks, which have not been systematically examined before. OCRGenBench addresses the critical void by providing the most comprehensive, unified testbed for evaluating generative models' OCR generative skills across diverse scenarios. We hope it establishes a standard benchmark for OCR generative evaluation, guiding future research toward more capable and reliable OCR generative systems.

Our main contributions are summarized as follows:
\begin{itemize}
	\item We propose OCRGenBench, a novel benchmark designed for holistic evaluation of generation models' OCR generative capabilities. OCRGenBench unifies five text categories and OCR generative 33 tasks with 1,060 manually annotated, challenging samples, establishing the most comprehensive and diverse benchmark to date.
	\item We propose OCRGenScore, a unified metric that evaluates text accuracy, structural consistency, and instruction following across diverse OCR generative tasks. Coupled with OCRGenBench, it enables fully automated evaluation of OCR generative performance and facilitates straightforward model comparisons.
	\item We systematically evaluate 19 SOTA generative models, encompassing both specialized and unified, closed-source and open-source models, revealing that most scores are below 60. Through in-depth analysis across tasks and text categories, we identify critical bottlenecks of current generative models that have been inadequately captured by existing benchmarks and distill promising directions for future studies.
\end{itemize}

\vspace{-3pt}
\section{Related Work}
\vspace{-5pt}
\subsection{State-of-the-Art Generative Model}
\vspace{-3pt}
Mainstream generation approaches fall into two types: specialized image generation models and unified understanding and generation models. The former targets merely image generation or editing, while the latter is capable of both understanding images and generating them.

\subsubsection{Specialized Generation Model}
Current specialized generation models are primarily built upon diffusion or flow matching frameworks, due to their remarkable image generation capabilities. Earlier methods leverage the diffusion architecture, such as the DALL$\cdot$E series \cite{dalle2021,dalle22022,dalle32023}, Imagen \cite{imagen2022}, and Stable Diffusion series \cite{ldm2022cvpr,sdxl2023podell,sd32024esser}, pioneering controllable text-to-image (T2I) image generation. Subsequent works advance controllability through additional conditions \cite{controlnet2023zhang,t2iadapter2024mou}, subject-driven synthesis \cite{ipadapter2023ye,omnicontrol2025tan}, and text-guided editing \cite{glide2022icml,instructpix2pix2023cvpr,emuedit2024}. Recently, flow matching-based methods \cite{fluxkontext2025,flux22025,omnicontrol2025tan}, which are often combined with diffusion transformer (DiT) \cite{dit2023iccv}, have gained popularity. Beyond separate solutions, a growing body of research \cite{seedream2025seedream,qwenimage2025wu,omnigen2025cvpr,omnigen22025wu,flux22025,flux2klein2026,longcat2025meituan} has shifted focus to unified image generation, \emph{i.e.}, performing T2I generation, image editing, and other generation tasks with one model, representing a new frontier for better scalability and efficiency.

\subsubsection{Unified Understanding and Generation Model}
Beyond single-modality generation, the field has evolved toward multimodal generation, yielding models capable of generating both textual and visual content. This line of approaches, termed unified understanding and generation models, effectively facilitates image generation with their context understanding skills or vice versa. Typically built upon multimodal large language models (MLLMs) with autoregressive objectives, they fall into two categories: external tool-assisted and single Transformer-based approaches. The first type utilizes an external diffusion model (\emph{e.g.}, Stable Diffusion) behind MLLM as a decoder for image generation \cite{emu2023sun,emu22024sun,nextgpt2024,seedx2025ge,metamorph2024tong,metaquries2025}. For example, Emu \cite{emu2023sun} and Emu2 \cite{emu22024sun} input the LLM's output as conditions to the diffusion model to generate images.

To achieve more efficient architectures and enhance synergy between understanding and generation, researchers seek to leverage a single transformer to unify both capabilities. These approaches can be divided into three meta-categories. (1) Pure autoregressive models. They use VQGAN or VQVAE to tokenize images into discrete tokens for multimodal autoregressive learning, and detokenize the output features back into images \cite{unifiedio22024lu,chameleon2024,lwm2025liu,janus2025wu,emu32024wang,janus4o2025chen}. (2) Diffusion/flow matching-embedded models. Beyond only autoregressive modeling, this line of work combines diffusion/flow matching loss with next-token prediction (NTP) loss to improve image generation \cite{transfusion2024zhou,showo2025xie,showo22025,janusflow2025ma,blip3o2025chen,uniworld2025lin,bagel2025deng,mogao2025liao,emma2025he,tuna2025liu}. They can switch roles between language models for autoregressive text generation and diffusion models for parallel image synthesis. Given flow matching's effectiveness, an increasing number of models adopt this technique for better performance \cite{janusflow2025ma,showo22025,emma2025he,tuna2025liu}. (3) Models with lightweight diffusion heads. Unlike the second meta-category, these approaches retain only NTP loss for training while incorporating an additional, lightweight diffusion head \cite{mar2024li} alongside the MLLM, which is conditioned on the MLLM's hidden states for image generation \cite{unifluid2025fan,sun2024multimodal,harmon2025wu}.

Early general-domain models neglect the importance of text synthesis, spurring the development of generation models specific to text synthesis \cite{glyphcontrol2023nips,anytext2024tuo,dreamtext2025wang} that sacrifice the generation abilities of natural objects. Recent models transcend this trade-off, simultaneously achieving strong natural scene synthesis and excellent text generation and editing performance, exemplified by Nano Banana Pro \cite{nanobananapro2025} and Qwen-Image \cite{qwenimage2025wu}. This reflects an emerging trend where SOTA models increasingly prioritize text rendering as a foundational capability rather than delegating it to specialized solutions, advancing this field toward more practical and intelligent generation.

\begin{table*}[t]
	\renewcommand{\arraystretch}{1.25}
	\centering
	\caption{Comparison between OCRGenBench and other visual text synthesis benchmarks. \emph{Doc.} denotes document, \emph{Hand.} denotes handwriting, \emph{Scene.} denotes scene text, \emph{Art.} denotes artistic text, and \emph{Lay.} denotes layout-rich text. \textbf{I2I Task} refers to OCR image-to-image translation tasks like document dewarping. \textbf{New Data} means whether this dataset contains new data beyond purely aggregating existing datasets.}
	\label{Table::dataset_comp}
	\resizebox{\textwidth}{!}{
		\begin{tabular}{l c c c c c c c c c c}
			\toprule
			\multirow{2}{*}{\textbf{Benchmark}} & \multicolumn{2}{c}{\textbf{Text Category}} & \multicolumn{4}{c}{\textbf{OCR Generative Task}} & \multirow{2}{*}{\textbf{New Data}} & \multirow{2}{*}{\textbf{Multilingual}} & \multirow{2}{*}{\textbf{Human Annotation}} & \multirow{2}{*}{\textbf{Complexity}}\\
			\cmidrule(r){2-3}\cmidrule(r){4-7}
			~ & \textbf{Scenario} & \textbf{\#} & \textbf{T2I Generation} & \textbf{Text Editing} & \textbf{I2I Task*} & \textbf{\#} & ~ & ~ & ~ & ~\\
			\hline
			LAION-Glyph \cite{glyphcontrol2023nips} & \emph{Scene.}, \emph{Lay.} & 2 & \textcolor{darkgreen}{\ding{51}} & \textcolor{deepred}{\ding{56}} & \textcolor{deepred}{\ding{56}} & 1 & \textcolor{darkgreen}{\ding{51}} & \textcolor{deepred}{\ding{56}} & \textcolor{deepred}{\ding{56}} & simple\\
			AnyText-Benchmark \cite{anytext2024tuo} & \emph{Scene.}, \emph{Lay.} & 2 & \textcolor{darkgreen}{\ding{51}} & \textcolor{deepred}{\ding{56}} & \textcolor{deepred}{\ding{56}} & 1 & \textcolor{deepred}{\ding{56}} & \textcolor{darkgreen}{\ding{51}} & \textcolor{deepred}{\ding{56}} & simple\\
			CVTG-2K \cite{cvtg2k2025} & \emph{Scene}, \emph{Doc.}, \emph{Lay.} & 3 & \textcolor{darkgreen}{\ding{51}} & \textcolor{deepred}{\ding{56}} & \textcolor{deepred}{\ding{56}} & 1 & \textcolor{darkgreen}{\ding{51}} & \textcolor{deepred}{\ding{56}} & \textcolor{deepred}{\ding{56}} & diverse\\
			Lex-Bench \cite{zhao2025lex} &  \emph{Scene.}, \emph{Lay.}, \emph{Art.} & 3 & \textcolor{darkgreen}{\ding{51}} & \textcolor{deepred}{\ding{56}} & \textcolor{deepred}{\ding{56}} & 1 & \textcolor{darkgreen}{\ding{51}} & \textcolor{deepred}{\ding{56}} & \textcolor{deepred}{\ding{56}} & diverse\\
			TextEditBench \cite{texteditbench2025gui} & \emph{Doc.}, \emph{Art.}, \emph{Lay.} & 3 & \textcolor{deepred}{\ding{56}} & \textcolor{darkgreen}{\ding{51}} & \textcolor{deepred}{\ding{56}} & 1 & \textcolor{darkgreen}{\ding{51}} & \textcolor{darkgreen}{\ding{51}} & \textcolor{darkgreen}{\ding{51}} & diverse\\
			VTPBench \cite{vtpbench2025} & \emph{Doc.}, \emph{Scene.} & 2 & \textcolor{darkgreen}{\ding{51}} & \textcolor{darkgreen}{\ding{51}} & \textcolor{darkgreen}{\ding{51}} & 6 & \textcolor{deepred}{\ding{56}} & \textcolor{darkgreen}{\ding{51}} & \textcolor{deepred}{\ding{56}} & simple\\
			\midrule
			OCRGenBench (\textbf{Ours}) & \emph{Doc.}, \emph{Hand.}, \emph{Scene.}, \emph{Art.}, \emph{Lay.} & \textbf{5} & \textcolor{darkgreen}{\ding{51}} & \textcolor{darkgreen}{\ding{51}} & \textcolor{darkgreen}{\ding{51}} & \textbf{33} & \textcolor{darkgreen}{\ding{51}} & \textcolor{darkgreen}{\ding{51}} & \textcolor{darkgreen}{\ding{51}} & \textbf{diverse}\\
			\noalign{\vspace{-2pt}}
			\bottomrule
	\end{tabular}}
\end{table*}

\subsection{Benchmark for Visual Text Generation and Editing}
\label{sec::benchmark_visual_text}
While existing image generation/editing benchmarks mostly target natural scenes \cite{geneval2023,hu2024ella,liu2025step1x,imgedit2025ye,oneigbench2025,longtext2025geng}, few incorporate visual text samples, and those that do remain quite limited. For example, DrawBench \cite{imagen2022} contains only 21 text rendering prompts out of 200 prompts; ImgEdit-Bench \cite{imgedit2025ye} contains 12 visual text editing samples among 734 samples. Some works put forward benchmarks \cite{anytext2024tuo,cvtg2k2025,zhao2025lex,texteditbench2025gui,vtpbench2025} dedicated to evaluating visual text generation or editing have been proposed. For T2I generation, AnyText-Benchmark \cite{anytext2024tuo} comprises 2,000 prompt-image pairs evaluating multilingual textual content generation, while CVTG-2K \cite{cvtg2k2025} provides 2,000 prompts for complex scenarios involving multi-region text, long text, and varied styles. Regarding text editing, TextEditBench \cite{texteditbench2025gui} evaluates the faithfulness of visual text editing in reasoning-intensive scenarios that require models to understand physical plausibility, linguistic meaning, \emph{etc.} Beyond traditional T2I generation/text editing, VTPBench \cite{vtpbench2025} separately examines six OCR generative tasks, such as scene text editing, generation, and removal. However, its coverage of text content and tasks remains limited.

Table~\ref{Table::dataset_comp} presents a multi-dimensional comparison between current visual text synthesis benchmarks and OCRGenBench. Despite these established benchmarks, they predominantly focus on either T2I generation or text editing, neglecting other OCR generative tasks critical for assessing models' text synthesis abilities. In addition, they are limited in text categories, typically confined to scene text and posters, and present insufficient difficulty and complexity, particularly in scenarios involving extensive, small, and dense text. Hence, there is a pressing need for a unified, comprehensive benchmark to evaluate generative models' text synthesis performance from an OCR-specialized perspective, which drives us to develop our OCRGenBench dataset.

%Optical character recognition (OCR) is a long-standing domain that involves extracting, processing, and understanding of visual text content. Due to its close relationship to visual perception and understanding, OCR abilities, \emph{e.g.}, visual question answering on documents, text location and recognition, are often incorporated in the evaluation of multimodal large language models (MLLMs). This leads to the development of many OCR-related benchmarks aiming for perceptual and understanding testing. Initial attempts target separate domains, including DocVQA \cite{docvqa2021wacv} for documents, ChartVQA \cite{chartqa2022masry} for charts, \emph{etc.} Recent advancements \cite{ocrbench2024liu,ccocr2025iccv,ocrbenchv22024,omnidocbench2025cvpr,ocrreasoning2025} consider the evaluation of multiple and compounded scenarios. OCRBench \cite{ocrbench2024liu}, OCRBench v2 \cite{ocrbenchv22024}, CC-OCR \cite{ccocr2025iccv} assess MLLM's text perceptual abilities across multiple OCR tasks and scenarios. OmniDocBench \cite{omnidocbench2025cvpr} provides a flexible and comprehensive evaluation on the parsing of multiple PDF document types. OCR-Reasoning \cite{ocrreasoning2025} benchmarks MLLMs' complex reasoning capabilities in text-rich visual understanding scenarios.

\begin{figure*}[t]
	\centering
	\includegraphics[width=\textwidth]{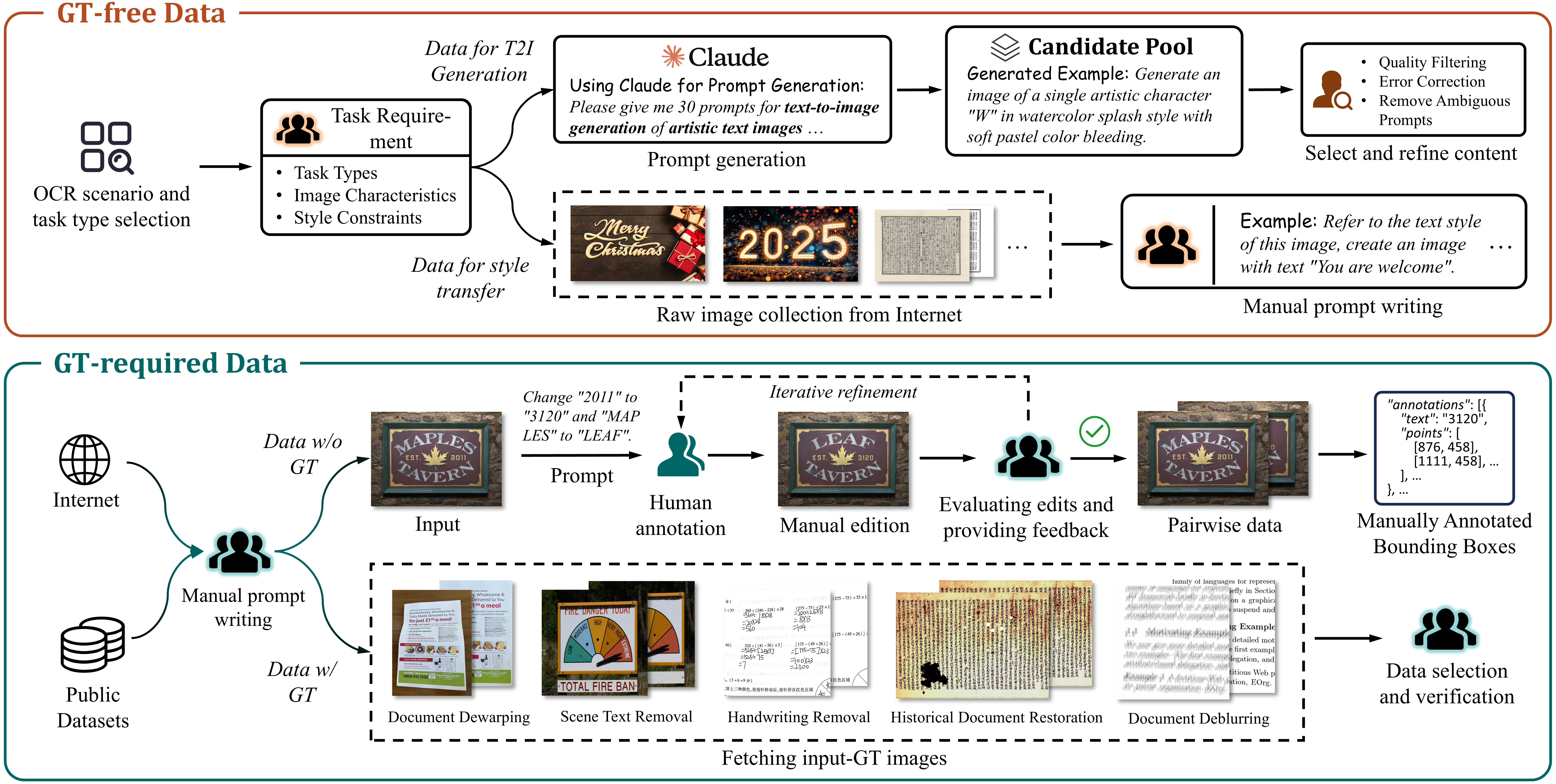}
	\caption{Data construction workflow of OCRGenBench.}
	\label{Fig::workflow}
\end{figure*}

\begin{figure*}[t]
	\centering
	\includegraphics[width=\textwidth]{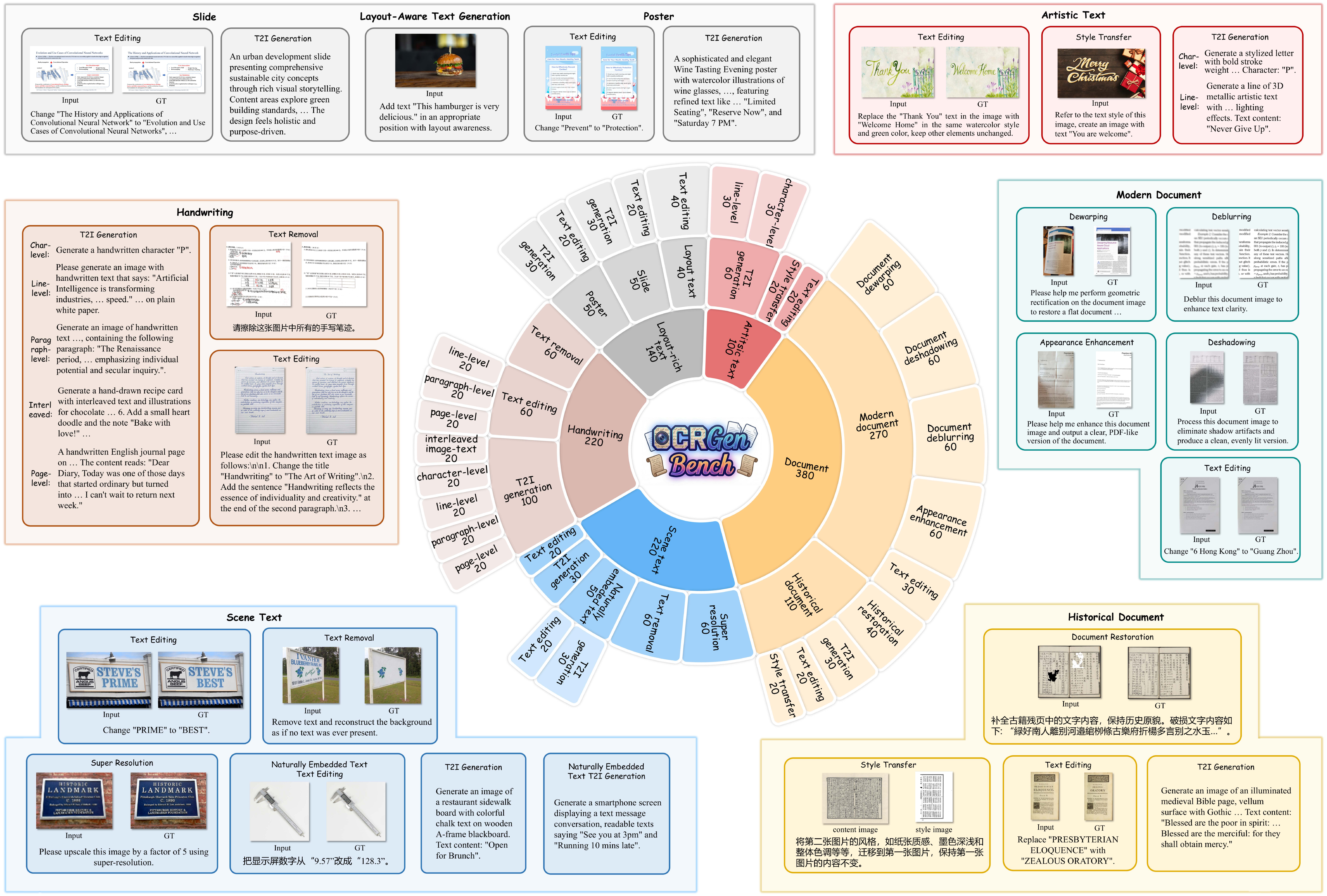}
	\caption{Data quantity distribution and examples of OCRGenBench.}
	\label{Fig::example}
\end{figure*}

\section{OCRGenBench Benchmark}
In this section, we first elaborate on the definition of text categories and OCR generative tasks covered by OCRGenBench, and then illustrate the data construction process and data statistics. Finally, we discuss the meaning and significance of OCRGenBench.

\subsection{Text Scenario and Task Description}
To enable a comprehensive coverage of text categories, OCRGenBench incorporates five major text types, including: document, handwriting, scene text, artistic text, and layout-rich text. Each category encompasses multiple OCR generative tasks targeting specific aspects of visual text synthesis, resulting in a total of 33 tasks. The definitions of these scenarios and tasks are detailed as follows.

\subsubsection{Text Category}
\textbf{Document}. Document images include both modern and historical documents. Modern documents are essential for digital-era information management, while historical documents preserve valuable cultural heritage. Both types are characterized by large amounts, dense, and variable-sized text and complex layouts, whose preprocessing and digitization present significant challenges for generative models.

\textbf{Handwriting}. Handwriting is ubiquitous in daily life, such as personal notes, letters, and exam papers. Handwriting generation \cite{notestylizedgen2025cvpr} enables writing assistance \cite{ha2018a,sketch2process2023} and data augmentation for handwriting analysis \cite{luo2019moran,msds2022zhang,pavenet2025zhang}, while text editing/removal facilitates privacy protection and educational applications \cite{li2024scene,huang2023ensexam}. However, replicating diverse writing styles and managing large character volumes with extreme aspect ratios during image generation remain challenging for generative models.

\textbf{Scene Text}. Scene text refers to textual information appearing naturally in real-world environments such as license plates and street signs. In contrast to documents or handwriting, scene text exhibits unconstrained visual properties including curved layouts \cite{liu2020abcnet}, varied orientations \cite{east2017zhou}, and perspective distortions \cite{aster2019shi}.

\textbf{Artistic Text}. Artistic text is stylized typography incorporating creative graphical elements or decorative fonts to achieve aesthetic appeal beyond standard formatting. The generation of artistic text is gaining increasing popularity across advertising, entertainment, and social media \cite{bai2024intelligent}, driving the advances of related models.

\textbf{Layout-Rich Text}. Layout-rich text includes slides and posters, integrating text, graphics, and intricate layouts for visual communication in various scenarios, such as marketing and education. Unlike artistic text that requires primarily visual aesthetics, their generation/editing requires understanding both textual semantics, design principles, and layout arrangement, driving us to incorporate them into the assessment.

\subsubsection{OCR Generative Task}
\textbf{Text-to-Image (T2I) Generation ($\times$12)}.  This task requires models to generate images from textual instructions. We prompt models to generate different types and levels (line, page, \emph{etc.}) of content. ``$\times$12'' indicates T2I generation appears 12 times in different text scenarios.

\textbf{Text Editing ($\times$10)}. This task modifies specific textual content within images according to instructions, while preserving other elements unchanged.

\textbf{Document Dewarping}. This task corrects geometric distortions of curved, warped, or folded document images. Models are required to output a flattened document surface with intact text \cite{docaligner2026zhang}.

\textbf{Document Deshadowing}. This task requires the models to remove shadows from document images \cite{deshadow2023iccv}.

\textbf{Document Deblurring}. This task refers to removing blur artifacts from document images to restore sharp, readable text and clear visual content \cite{degan2022souibgui}.

\textbf{Document Appearance Enhancement}. This task, also known as document illumination rectification \cite{doctr2021feng}, corrects uneven lighting and removes degradations (shadows, bleed-through) in camera-captured documents.

\textbf{Historical Document Restoration}. This task \cite{historical2013} recovers deteriorated historical documents to restore readability. With missing content provided in instructions, models must generate text characters matching the original writing style along with visually coherent backgrounds \cite{autohdr2025zhang}.

\textbf{Historical Document Style Transfer}. This task transfers the writing and background style of the second historical document to the first one, while preserving the original text content.

\textbf{Handwriting Removal}. This task removes all handwritten text from images.

\textbf{Scene Text Removal}. This task erases textual content of any writing style from scene images.

\textbf{Scene Text Super Resolution}. This task enhances the spatial resolution of scene text images while preserving textual content and visual fidelity.

\textbf{Artistic Text Style Transfer}. This task generates a new image with specific text content that mimics the reference image's style. Unlike historical document style transfer, this task requires generating target content rather than preserving original content.

\textbf{Layout-Aware Text Generation}. This task, also known as content-aware layout generation \cite{contentaware2024cvpr}, requires placing text naturally within images without obliging the original graphical component.

% T2I, Editing, T2I * 11, Editing * 9
% Document Dewarping, deshadowing, deblurring, enhancement
% historical restoration, style transfer
% scene text removal, scene text super resolution
% handwritten text removal
% artistic text style transfer
% layout-aware text generation

\begin{figure*}[t]
	\centering
	\includegraphics[width=\textwidth]{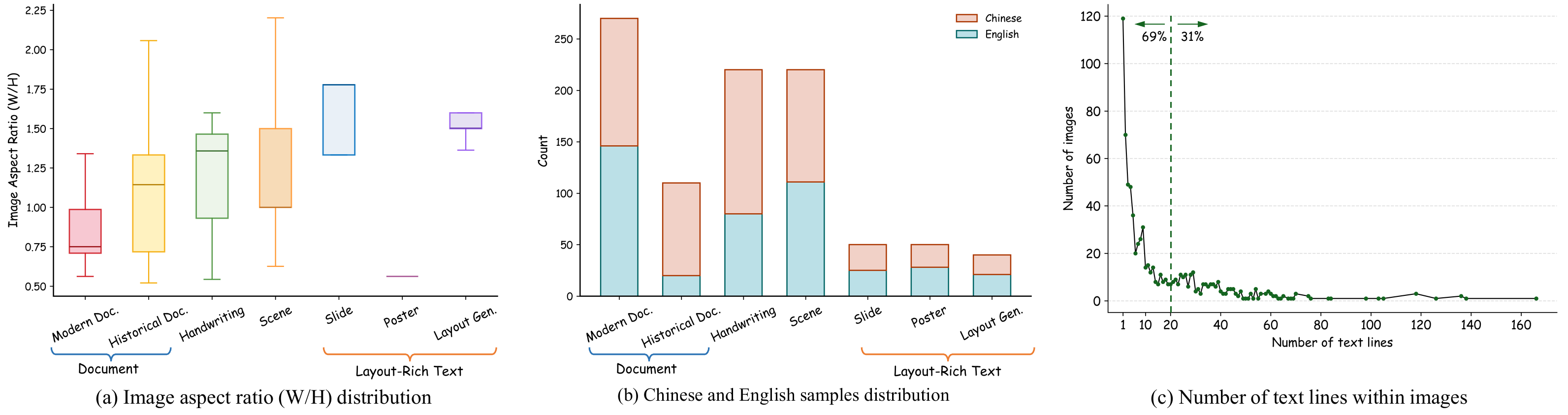}
	\caption{Distribution statistics of OCRGenBench. Doc. denotes Document. Gen. denotes Generation.}
	\label{Fig::stat}
\end{figure*}

\subsection{Data Construction}
\label{sec::data_construct}
OCRGenBench adapts to the instruction-based paradigm of modern generative models, where each sample comprises a text prompt with optional input and ground truth (GT) images. Fig.~\ref{Fig::workflow} illustrates our data construction pipeline. We categorize the 33 tasks into two types based on evaluation requirements: GT-free tasks, which involve open-ended generation without definitive answers (\emph{e.g.}, T2I generation, artistic style transfer), and GT-required tasks, which encompass editing operations with verifiable outputs (e.g., text editing, text removal).

\textbf{GT-free data construction.} GT-free tasks include T2I generation (requiring only prompts) and artistic/historical document style transfer (requiring prompts and input images). For T2I generation, we employ Claude-Sonnet-4.5 \cite{claude452025} to generate diverse prompts by specifying task types and image characteristics. For instance, to create prompts for artistic text generation, we instruct Claude by: ``Please generate 30 prompts for \underline{text-to-image generation} of \underline{artistic text images}. Specify the text content and ensure diverse artistic styles.'' The generated prompts span both English and Chinese to enable multilingual evaluation. We manually curate high-quality prompts from the candidates and correct erroneous content to ensure accuracy. As for style transfer tasks, we collect raw images from the Internet and manually write corresponding prompts.

\textbf{GT-required data construction.} GT-required data construction follows a two-stage approach depending on GT availability. For tasks with existing GT images (\emph{e.g.}, document dewarping, scene text removal), we collect data from eight public datasets \cite{icdar2019sign,erasenet2020liu,scenetext2020,ocrbenchv22024}. We manually compose task-specific prompts and pair them with corresponding input-GT image pairs, followed by quality verification.

For text editing tasks lacking GT images, we implement a rigorous human-in-the-loop annotation process. We first design editing prompts with specific modifications. Then, two professional image editors are hired to create edited images following these instructions. The resulting images undergo iterative refinement through at least two rounds of author review and editor revision. After the editing is finished, we annotate the spatial coordinates and textual content of edited regions to facilitate automated evaluation. This manual annotation strategy ensures data originality and prevents training contamination, distinguishing OCRGenBench from existing image generation datasets and benchmarks. All GT-required data includes both Chinese and English samples with language-consistent prompts for comprehensive multilingual assessment. Some data examples are visualized in Fig.~\ref{Fig::example}.

\begin{table*}[t]
	\renewcommand{\arraystretch}{1.15}
	\centering
	\caption{Discrete statistics of OCRGenBench. Avg. $L_{Prompt}$ denotes the average prompt length.}
	\label{Table::stat}
	\resizebox{\textwidth}{!}{
		\begin{tabular}{l c c c c c}
			\toprule
			\multirow{2}{*}{\textbf{Text Category}} & \multicolumn{2}{c}{\textbf{Image Aspect Ratio (W/H})} & \multirow{2}{*}{\textbf{Resolution Range}} & \multirow{2}{*}{\textbf{Avg. $L_{Prompt}$ (EN)}} & \multirow{2}{*}{\textbf{Avg. $L_{Prompt}$ (ZH)}}\\
			\cmidrule(r){2-3}
			~ & \textbf{Minima} & \textbf{Maxima} & ~ & ~ & ~\\
			\hline
			Modern Document (\emph{Document}) & 0.578:1 & 1.950:1 & 300$\times$300 - 4,958$\times$6,985 & 908 & 832\\
			Historical Document (\emph{Document}) & 0.521:1 & 1.678:1 & 300$\times$261 - 6,303$\times$4,960 & 237 & 9397\\
			Scene Text & 0.626:1 & 2.159:1 & 270$\times$360 - 2,448$\times$3,264 & 494 & 807\\
			Handwriting & 0.544:1 & 16.719:1 & 453$\times$64 - 2,160$\times$2,910 & 2567 & 3780\\
			Artistic Text & 1.339:1 & 5.838:1 & 566$\times$120 - 2,114$\times$876 & 538 & 796\\
			Poster (\emph{Layout-Rich Text}) & 0.497:1 & 0.635:1 & 540$\times$960 - 1440$\times$2560 & 655 & 625\\
			Slide (\emph{Layout-Rich Text}) & 1.333:1 & 1.778:1 & 2560$\times$1440 - 10,250$\times$5,760 & 1729 & 3007\\
			Layout-Aware Text (\emph{Layout-Rich Text}) & 1.473:1 & 1.600:1 & 1,024$\times$682 - 5,495$\times$3,440 & 152 & 255\\
			\noalign{\vspace{-2pt}}
			\bottomrule
	\end{tabular}}
\end{table*}

\subsection{Data Statistics}
\label{sec::statistics}
In total, OCRGenBench comprises 1,060 samples spanning five text categories and 33 OCR generative tasks. Comprehensive statistics are presented in Fig.~\ref{Fig::example}, Fig.~\ref{Fig::stat} and Table~\ref{Table::stat}, respectively.

\textbf{Task and scenario distribution}. Fig.~\ref{Fig::example} illustrates the sample quantity distribution across different task types and scenarios, following the hierarchical taxonomy in Fig.~\ref{Fig::mindmap}. This distribution ensures not only comprehensive coverage of generation and editing capabilities across multiple text modalities, but also sufficient samples within each task and category for reliable evaluation.

\textbf{Aspect ratio and resolution diversity}. Unlike existing datasets dominated by square or near-square images, OCRGenBench intentionally incorporates heterogeneous aspect ratios to reflect real-world conditions. Fig.~\ref{Fig::stat} (a) demonstrates distinct aspect ratio distributions across text categories. Table~\ref{Table::stat} reveals extreme cases, such as line-level handwriting with ratios reaching 16.719:1. Image resolutions vary substantially, ranging from 453$\times$64 to 10,250$\times$5,760 pixels. This heterogeneity challenges models to preserve original aspect ratios and textual integrity across varying dimensions, enabling realistic evaluation under authentic layout conditions.

\textbf{Multilingual composition}. Fig.~\ref{Fig::stat} (b) presents the Chinese-English distribution across text categories. Most categories maintain balanced 50:50 ratios, with exceptions in historical documents and handwriting due to data availability constraints. This bilingual design facilitates comprehensive multilingual assessments.

\textbf{Text density and prompt complexity}. Fig.~\ref{Fig::stat} (c) shows text line distributions: approximately 69\% of images contain fewer than 20 lines (minimum one line), while the remaining 31\% exhibit substantially higher text density. This variation enables evaluation across sparse and dense text scenarios. Furthermore, average prompt lengths vary significantly by text type, where slide generation requires detailed specifications with longer instructions, while scene text generation typically requires one or two concise sentences. This diversity reflects the varying complexity and contextual requirements across different text generation tasks.

%Overall, the coverage comprehensiveness, resolution diversity, text density, plus with large amount text generation requirement in some page-level content, makes OCRGenBench a unified, comprehensive, and challenging benchmark for evaluating OCR generative performance.

\subsection{Discussion: The Significance of OCRGenBench}
\textbf{Constraints of existing benchmarks}. As shown in Table~\ref{Table::dataset_comp}, existing benchmarks remain fragmented, exhibiting limited text scenarios (only scene text, poster, or document), restricted task coverage (primarily T2I generation or text editing, neglecting OCR generative tasks), and insufficient difficulty. This compartmentalization creates critical blind spots. First, isolated text categories mask cross-domain vulnerabilities. A model excelling at printed documents may fail on handwritten or artistic content. Second, single-task evaluation cannot reveal whether performance transfers across different operations, where models may generate legible scene text yet fail at document editing. Third, the absence of challenging samples, such as those featuring dense text, extreme aspect ratios, or complex style preservation, leads to inflated metrics that do not reflect a model's true real-world reliability.

\textbf{Necessity for an OCR generative benchmark}. With natural image synthesis reaching photorealistic quality, the more challenging textual content synthesis has emerged as the next frontier. Recently, an increasing number of models claim robust text generation capabilities. Yet as discussed above, current benchmarks fall short in task and text categories coverage, preventing comprehensive, standardized evaluation of models' true OCR generative abilities. OCRGenBench addresses this gap by providing the most comprehensive and unified benchmark (see Table~\ref{Table::dataset_comp}), enabling accurate, holistic assessment of generative models' text synthesis capabilities.

\textbf{Comprehensiveness, unity, specialty, and diversity}. OCRGenBench includes five common text image categories, which can be further divided into eight fine-grained types (modern documents, historical documents, handwriting, scene text, posters, slides, and layout-aware images), and 33 OCR generative tasks. Unlike existing benchmarks that evaluate isolated subsets, OCRGenBench unifies multiple texts and tasks within a single framework, enabling holistic assessment across the full spectrum of text synthesis capabilities. Some tasks, such as document dewarping and historical document restoration, are technically sophisticated, requiring specialized domain knowledge beyond general synthesis. Furthermore, as described in Sec.~\ref{sec::statistics}, the benchmark exhibits substantial diversity in image aspect ratios, resolutions, text densities, generated text length (from single-character to entire pages with 1300+ characters), and bilingual content, providing realistic, challenging conditions for evaluations.

\section{OCRGenScore Metric}
To comprehensively assess generation quality across 33 OCR generative tasks, we introduce OCRGenScore, a composite metric that combines multiple evaluation approaches tailored to different task characteristics. The metric computation details are described below.

\subsection{Detailed Metric Computation}
\textbf{T2I Generation}. We use VIEScore \cite{viescore2024ku} and Accurate Rate (AR) to assess the generated image's quality, which are formulated as:
\begin{equation}
	\label{eq::viescore}
	[S_{SC},S_{PQ}] = VIEScore(image,prompt); S_{O} = \sqrt{S_{SC} \times S_{PQ}},
\end{equation}
\begin{equation}
	\label{eq::ar}
	AR = (N_t - D_e - S_e - I_e) / N_t.
\end{equation}
Since T2I generation is an open-ended task with unlimited valid solutions, it is infeasible to use definitive ground truths for metric computation. Thus, we use VIEScore, an LLM-as-Judge scheme to assess generated images' prompt semantic consistency $S_{SC}$ and perceptual quality (\emph{e.g.}, visual naturalness) $S_{PQ}$ on [0,10] scales, with overall score $S_{O}$ computed as in Eq.~\ref{eq::viescore}. In addition, we use the accuracy rate (AR) to evaluate the correctness of the generated text, formulated in Eq.~\ref{eq::ar}. The textual content is detected and recognized using PPOCRv5 \cite{ppocr2020du}, and then compared against the content specified in the prompt (as GT). In Eq.~\ref{eq::ar}, $N_t$ is the total number of characters in the annotations, while $D_e$, $S_e$, and $I_e$ denote deletion, substitution, and insertion errors, respectively.

\begin{table*}[t]
	\renewcommand{\arraystretch}{1.2}
	\centering
	\caption{Overview of evaluated models. \emph{Pub. Date} denotes publication date. \emph{Und.} denotes understanding. \emph{Gen.} denotes Generation. \emph{GPU Memory} specifically refers to GPU memory requirements for model inference.}
	\resizebox{\linewidth}{!}{
		\begin{tabular}{l c c c c c c}
			\toprule
			\noalign{\vspace{-1pt}}
			\textbf{Model} & \textbf{Pub. Date} &\textbf{Type} & \textbf{Availability} & \textbf{\#Parameters} & \textbf{Max Context Length} & \textbf{GPU Memory}\\
			\hline
			Nano Banana Pro \cite{nanobananapro2025} & 2025.11.20 & \textcolor{purple}{Unified Und. \& Gen.} & Closed-Source & - & - & -\\
			GPT Image 1.5 \cite{gptimage152025} & 2025.3.25 & \textcolor{deepred}{Specialized Generation} & Closed-Source & - & - & -\\
			Seedream 4.5 \cite{seedream2025seedream} & 2025.12.3 & \textcolor{deepred}{Specialized Generation} & Closed-Source & - & - & -\\
			BAGEL \cite{bagel2025deng} & 2025.5.20 & \textcolor{purple}{Unified Und. \& Gen.} & Open-Source & 14B & 32,768 tokens & 2 $\times$ 48GB\\
			OmniGen2 \cite{omnigen22025wu} & 2025.6.16 & \textcolor{purple}{Unified Und. \& Gen.} & Open-Source & 7B & 128,000 tokens & 48 GB\\
			InternVL-U \cite{internvlu2026tian} & 2026.3.11 & \textcolor{purple}{Unified Und. \& Gen.} & Open-Source & 4B & 40,960 tokens & 48 GB\\
			Show-o2 \cite{showo22025} & 2025.6.18 & \textcolor{purple}{Unified Und. \& Gen.} & Open-Source & 7B & 1024 tokens & 48 GB\\
			ILLUME+ \cite{illumeplus2025huang} & 2025.4.2 & \textcolor{purple}{Unified Und. \& Gen.} & Open-Source & 7B & 32,768 tokens & 48 GB\\
			Janus-4o \cite{janus4o2025chen} & 2025.6.22 & \textcolor{purple}{Unified Und. \& Gen.} & Open-Source & 7B & 4,096 tokens & 48 GB\\
			Qwen-Image-Edit-2511 \cite{qwenimage2025wu} & 2025.12.25 & \textcolor{deepred}{Specialized Generation} & Open-Source & 20B & 128,000 tokens & 2 $\times$ 48 GB\\
			LongCat-Image \cite{longcat2025meituan} & 2025.12.8 & \textcolor{deepred}{Specialized Generation} & Open-Source & 6B & 128,000 tokens & 48 GB\\
			Stable Diffusion 3.5-Large \cite{sd32024esser} & 2024.10.22 & \textcolor{deepred}{Specialized Generation} & Open-Source & 8B & 77 tokens & 48 GB\\
			FlUX.1 Kontext-dev \cite{fluxkontext2025} & 2025.6.26 & \textcolor{deepred}{Specialized Generation} & Open-Source & 12B & 77 tokens & 48 GB\\
			FlUX.2-dev \cite{fluxkontext2025} & 2025.11.25 & \textcolor{deepred}{Specialized Generation} & Open-Source & 32B & 128,000 tokens & 4 $\times$ 48 GB\\
			Flux.2-Klein-9B \cite{flux2klein2026} & 2026.1.16 & \textcolor{deepred}{Specialized Generation} & Open-Source & 9B & 40,960 tokens & 48 GB\\
			Z-Image-Turbo \cite{zimage2025} & 2025.11.26 & \textcolor{deepred}{Specialized Generation} & Open-Source & 6B & 40,960 tokens & 48 GB\\
			Ovis-Image \cite{ovis2025wang} & 2025.11.29 & \textcolor{deepred}{Specialized Generation} & Open-Source & 7B & 40,960 tokens & 48 GB\\
			GLM-Image \cite{glmimage2026} & 2026.1.14 & \textcolor{deepred}{Specialized Generation} & Open-Source & 9B & 131,072 tokens & 48 GB\\
			FireRed-Image-Edit-1.1 \cite{firered2026} & 2026.3.8 & \textcolor{deepred}{Specialized Generation} & Open-Source & 20B & 128,000 tokens & 2 $\times$ 48 GB\\
			\noalign{\vspace{-1pt}}
			\bottomrule
	\end{tabular}}
	\label{Table::overview}
\end{table*}

\textbf{Text Editing}. We evaluate the accuracy of the edited text and its visual quality. Using position annotations of the edited text (described in Sec.~\ref{sec::data_construct}), we crop the same area in the generated image. We then extract the text content within using PPOCRv5 and compare it with the GT content to calculate AR. We also adopt LPIPS \cite{lpips2018cvpr} to calculate the visual resemblance and perceptual quality between the GT and edited images. 

\textbf{Document Dewarping}. To evaluate dewarping effectiveness, the community typically uses distortion-based metrics for their high accuracy. Accordingly, we adopt the Document-Distortion (DD) metric \cite{dd2025zhang} due to its superior performance and native Python implementation, which avoids the MATLAB dependencies of alternative metrics.

\textbf{Historical Document \& Artistic Style Transfer}. Both tasks are open-ended with multiple valid outputs, making reference-based quantitative assessment infeasible. We adapt VIEScore by modifying evaluation prompts to assess style composition, readability, and other relevant aspects. The overall score $S_O$ is computed as in Eq.~\ref{eq::viescore}. In addition, we compute AR in artistic style transfer to assess text accuracy.

\textbf{Historical Document Restoration}. Following common practices \cite{hdr2025yang,autohdr2025zhang} in this field, we use LPIPS to evaluate the restoration quality.

\textbf{Layout-Aware Text Generation}. This task is open-ended since text can be placed in any feasible place. Therefore, we adapt VIEScore with tailored evaluation prompts to evaluate generated text quality (naturalness and alignment), structural consistency, and layout integrity. We also compute AR to assess text accuracy.

\textbf{Other Tasks}. For remaining OCR generative tasks, such as scene text removal and document deshadowing, we adopt MS-SSIM \cite{msssim2003}, the standard metric for traditional image-to-image tasks in the OCR field.

\subsection{Normalization and Aggregation}
OCRGenScore aggregates all task-specific metrics into a unified score. We normalize each metric to [0,1] (higher is better): VIEScore $S_O$ is divided by 10; LPIPS is inverted to $1-\text{LPIPS}$; DD is transformed via $\text{DD}_\text{inv}=e^{-\text{DD}/10}$ to map $[0,+\infty]$ to $[0,1]$; MS-SSIM and AR are already in [0,1]. The final OCRGenScore is the mean of normalized metrics across different tasks, scaled by 100 for readability.

\section{Experiments}
\subsection{Baseline}
Our evaluation encompasses 19 models, spanning both closed-source and open-source domains. Closed-source models include Nano Banana Pro \cite{nanobananapro2025}, GPT Image 1.5 \cite{gptimage152025}, and Seedream 4.5 \cite{seedream2025seedream}, while open-source models include BAGEL \cite{bagel2025deng}, OmniGen2 \cite{omnigen22025wu}, InternVL-U \cite{internvlu2026tian}, Janus-4o \cite{janus4o2025chen}, ILLUME+ \cite{illumeplus2025huang}, Show-o2 \cite{showo22025}, Qwen-Image \cite{qwenimage2025wu}, LongCat-Image \cite{longcat2025meituan}, Stable Diffusion-3.5-Large \cite{sd32024esser}, Flux.1-dev \cite{fluxkontext2025}, Flux.2-dev \cite{flux22025}, Flux.2-Klein-9B \cite{flux2klein2026}, Z-Image \cite{zimage2025}, Ovis-Image \cite{ovis2025wang}, GLM-Image \cite{glmimage2026}, and FireRed-Image-Edit-v1.1 \cite{firered2026}. Nano-Banana-Pro, BAGEL, OmniGen2, InternVL-U, Janus-4o, ILLUME+, and Show-o2 are \textcolor{purple}{unified generation and understanding models}, while other models are \textcolor{deepred}{specialized generation models}. Note that models limited to T2I generation are evaluated exclusively on T2I tasks in OCRGenBench. An overview of these models is presented in Table~\ref{Table::overview}.

\begin{table*}[t]
	\renewcommand{\arraystretch}{1.2}
	\caption{Performance comparison on OCRGenBench across all tasks. {Doc.} denotes {Document}. {Appear. Enhance.} denotes Appearance Enhancement. {Hist. Doc. Rest.} refers to Historical Document Restoration. {Gen.} refers to Generation. The best results are marked in \textbf{bold} while the second-best are marked with \underline{underline}. Same notation throughout.}
	\begin{minipage}{\textwidth}
		\centering
		\resizebox{\linewidth}{!}{
			\begin{tabular}{l c cc cc c c c c}
				\toprule
				\multirow{2.2}{*}{\textbf{Model}} & \multirow{2.2}{*}{\textbf{OCRGenScore} $\uparrow$} & \multicolumn{2}{c}{\textbf{T2I}} & \multicolumn{2}{c}{\textbf{Editing}} & \textbf{Doc. Dewarping} &\textbf{Doc. Deshadowing} & \textbf{Doc. Deblurring} & \textbf{Appear. Enhance.}\\
				\cmidrule(lr){3-4}\cmidrule(lr){5-6}\cmidrule(lr){7-7}\cmidrule(lr){8-8}\cmidrule(lr){9-9}\cmidrule(lr){10-10}
				~ & ~ & \textbf{VIEScore} $\uparrow$ & \textbf{AR} $\uparrow$ & \textbf{1-LPIPS} $\uparrow$ & \textbf{AR} $\uparrow$ & \textbf{DD} $\uparrow$ & \textbf{MSSSIM} $\uparrow$ & \textbf{MSSSIM} $\uparrow$ & \textbf{MSSSIM} $\uparrow$\\
				\hline
				\multicolumn{10}{l}{\textbf{Closed-Source Model}}\\
				\hline
				\rowcolor{purple!10}
				\multicolumn{10}{l}{\textcolor{purple}{\emph{Unified Und. \& Gen. Model}}}\\
				Nano Banana Pro \cite{nanobananapro2025} & \textbf{77.19} & \underline{92.24} & \textbf{76.96} & 85.22 & \textbf{71.46} & \textbf{42.52} & \textbf{85.93} & \textbf{61.37} & \textbf{80.31}\\
				\rowcolor{deepred!10}
				\multicolumn{10}{l}{\textcolor{deepred}{\emph{Specialized Generation Model}}}\\
				GPT Image 1.5 \cite{gptimage152025} & 54.00 & \textbf{93.41} & 68.72 & 57.36 & 42.73 & 29.55 & 36.10 & 32.09 & 41.82\\
				Seedream 4.5 \cite{seedream2025seedream} & 63.35 & 90.45 & \underline{71.09} & 61.75 & 45.19 & 23.85 & 57.99 & 55.09 & 53.28\\
				\hline
				\multicolumn{10}{l}{\textbf{{Open-Source Model}}}\\
				\hline
				\rowcolor{purple!10}
				\multicolumn{10}{l}{\textcolor{purple}{\emph{Unified Und. \& Gen. Model}}}\\
				BAGEL \cite{bagel2025deng} & 59.11 & 57.08 & 14.95 & \textbf{87.03} & 15.07 & 20.80 & \underline{76.18} & 60.11 & \underline{76.27}\\
				OmniGen2 \cite{omnigen22025wu} & 54.24 & 59.96 & 26.37 & 65.83 & 13.82 & 21.21 & 64.26 & 43.68 & 65.17\\
				InternVL-U \cite{internvlu2026tian} & 43.64 & 66.25 & 44.17 & 53.87 & 17.21 & 28.69 & 49.45 & 30.80 & 46.83\\
				ILLUME+ \cite{illumeplus2025huang} & 28.39 & 43.15 & 5.46 & 42.71 & 2.68 & 27.21 & 42.78 & 40.85 & 38.28\\
				Janus-4o \cite{janus4o2025chen} & 29.58 & 53.59 & 13.63 & 45.53 & 5.80 & 24.88 & 34.28 & 37.12 & 39.33\\
				\rowcolor{deepred!10}
				\multicolumn{10}{l}{\textcolor{deepred}{\emph{Specialized Generation Model}}}\\
				Qwen-Image \cite{qwenimage2025wu} & 56.29 & 84.41 & 65.21 & 65.65 & 41.20 & 25.14 & 50.05 & 38.81 & 61.50\\
				LongCat-Image \cite{longcat2025meituan} & 66.39 & 84.02 & 67.51 & \underline{85.56} & \underline{54.53} & 28.04 & 72.12 & 60.05 & 56.62\\
				SD-3.5-Large \cite{sd32024esser} & 29.53 & 50.94 & 27.51 & 47.43 & 5.77 & 30.99 & 29.07 & 32.64 & 32.45\\
				Flux.1-Kontext-dev \cite{fluxkontext2025} & 36.51 & 39.58 & 21.69 & 53.76 & 15.13 & 24.30 & 30.36 & 30.80 & 40.90\\
				Flux.2-dev \cite{flux22025} & \underline{70.19} & 88.88 & 66.37 & 83.92 & 41.56 & \underline{41.41} & 67.97 & \underline{60.42} & 71.87\\
				Flux.2-Klein-9B \cite{flux2klein2026} & 59.28 & 82.84 & 39.63 & 81.18 & 31.10 & 24.44 & 57.98 & 51.35 & 56.74\\
				GLM-Image \cite{glmimage2026} & 50.12 & 83.53 & 69.36 & 70.97 & 21.54 & 24.99 & 43.15 & 41.13 & 51.59\\
				\noalign{\vspace{-1pt}}
				\bottomrule
		\end{tabular}}
	\end{minipage}
	\vfill
	\vspace{3pt}
	\begin{minipage}{\textwidth}
		\centering
		\resizebox{\linewidth}{!}{
			\begin{tabular}{l c c c c c c c c c c}
				\toprule
				\multirow{3.3}{*}{\textbf{Model}} & \multirow{3.3}{*}{\textbf{OCRGenScore} $\uparrow$} & \multicolumn{2}{c}{\textbf{Text Removal}} & \multicolumn{3}{c}{\textbf{Style Transfer}} & \multirow{2.2}{*}{\textbf{Hist. Doc. Rest.}} & \multirow{2.2}{*}{\textbf{Scene Text Super Resolution}} & \multicolumn{2}{c}{\multirow{2.2}{*}{\textbf{Layout-Aware Text Gen.}}}\\
				\cmidrule(lr){3-4}\cmidrule(lr){5-7}
				~ & ~ & \textbf{Handwriting} & \textbf{Scene Text} & \multicolumn{2}{c}{\textbf{Artistic Text}} & \textbf{Hist. Doc.} &&&&\\
				\cmidrule(lr){3-3}\cmidrule(lr){4-4}\cmidrule(lr){5-6}\cmidrule(lr){7-7}\cmidrule(lr){8-8}\cmidrule(lr){9-9}\cmidrule(lr){10-11}
				~ & ~ & \textbf{MSSSIM} $\uparrow$ & \textbf{MSSSIM} $\uparrow$ & \textbf{VIEScore} $\uparrow$ & \textbf{AR} $\uparrow$ & \textbf{VIEScore} $\uparrow$ & \textbf{1-LPIPS} $\uparrow$ & \textbf{MSSSIM} $\uparrow$ & \textbf{VIEScore} $\uparrow$ & \textbf{AR} $\uparrow$\\
				\hline
				\multicolumn{11}{l}{\textbf{Closed-Source Model}}\\
				\hline
				\rowcolor{purple!10}
				\multicolumn{11}{l}{\textcolor{purple}{\emph{Unified Und. \& Gen. Model}}}\\
				Nano Banana Pro \cite{nanobananapro2025} & \textbf{77.19} & \textbf{84.16} & \textbf{91.66} & 78.27 & 89.00 & 77.66 & \underline{74.15} & 64.67 & \textbf{88.87} & \textbf{100.00}\\
				\rowcolor{deepred!10}
				\multicolumn{11}{l}{\textcolor{deepred}{\emph{Specialized Generation Model}}}\\
				GPT Image 1.5 \cite{gptimage152025} & 54.00 & 47.59 & 51.38 & \textbf{87.16} & {94.22} & \textbf{80.57} & 46.18 & 23.23 & 85.69 & 97.77\\
				Seedream 4.5 \cite{seedream2025seedream} & 64.69 & 69.97 & 83.93 & \underline{81.52} & \textbf{99.34} & \underline{78.07} & 59.88 & 42.02 & 54.50 & {95.00}\\
				\hline
				\multicolumn{11}{l}{\textbf{Open-Source Model}}\\
				\hline
				\rowcolor{purple!10}
				\multicolumn{11}{l}{\textcolor{purple}{\emph{Unified Und. \& Gen. Model}}}\\
				BAGEL \cite{bagel2025deng} & 59.11 & \underline{80.23} & 79.78 & 33.19 & 22.80 & 44.49 & 73.71 & \textbf{87.67} & 75.31 & 32.99\\
				OmniGen2 \cite{omnigen22025wu} & 54.24 & 47.28 & 82.11 & 35.17 & 47.59 & 64.66 & 50.64 & \underline{83.75} & 72.58 & 43.33\\
				InternVL-U \cite{internvlu2026tian} & 43.64 & 57.19 & 61.13 & 53.41 & 59.99 & 20.75 & 37.45 & 17.74 & 53.94 & 85.83\\
				ILLUME+ \cite{illumeplus2025huang} & 28.39 & 48.93 & 68.42 & 0.50 & 2.26 & 1.00 & 25.84 & 21.88 & 7.67 & 3.29\\
				Janus-4o \cite{janus4o2025chen} & 29.58 & 43.69 & 35.77 & 25.00 & 8.17 & 16.53 & 28.60 & 19.05 & 41.55 & 17.32\\
				\rowcolor{deepred!10}
				\multicolumn{11}{l}{\textcolor{deepred}{\emph{Specialized Generation Model}}}\\
				Qwen-Image \cite{qwenimage2025wu} & 56.29 & 49.22 & 44.56 & 70.63 & \underline{95.47} & 72.33 & 59.48 & 27.65 & \underline{84.44} & \underline{98.93}\\
				LongCat-Image \cite{longcat2025meituan} & 66.39 & 78.11 & 89.76 & 69.22 & 94.27 & 69.74 & 66.86 & 24.37 & 83.57 & 96.14\\
				SD-3.5-Large \cite{sd32024esser} & 29.53 & 44.64 & 26.11 & 56.64 & 44.01 & 0.00 & 42.79 & 17.37 & 18.52 & 4.84\\
				Flux.1-Kontext-dev \cite{fluxkontext2025} & 36.51 & 38.34 & 31.87 & 58.66 & 52.59 & 52.90 & 43.24 & 22.49 & 49.00 & 28.52\\
				Flux.2-dev \cite{flux22025} & \underline{70.19} & 78.75 & 89.65 & 79.96 & 91.32 & 68.48 & \textbf{74.62} & 52.78 & {76.53} & 84.51\\
				Flux.2-Klein-9B \cite{flux2klein2026} & 59.28 & 65.71 & \underline{90.59} & 78.48 & 59.39 & 67.48 & 65.81 & 46.46 & 74.05 & 41.61\\
				GLM-Image \cite{glmimage2026} & 50.12 & 64.61 & 16.21 & 59.77 & 51.90 & 77.81 & 67.98 & 11.80 & 80.81 & 66.67\\
				\noalign{\vspace{-1pt}}
				\bottomrule
			\end{tabular}
		}
	\end{minipage}
	\label{Table::exp_across_all_task}
\end{table*}

\subsection{Implementation Details}
\label{sec::impl}
\textbf{Environment and Inference Configurations.} Experiments are conducted on NVIDIA RTX 6000 (48GB) GPUs, using PyTorch 2.6.0 \cite{pytorch2019paszke}, CUDA 11.8, and the HuggingFace \texttt{diffusers} and \texttt{transformers} \cite{huggingface2020wolf} libraries. Open-source models are loaded in \(\text{bfloat16}\) precision to optimize memory. Table~\ref{Table::overview} details the specific memory requirements per model. For open-source models, we adopt the hyperparameters (\emph{e.g.}, classifier-free guidance (CFG) scale and inference steps) explicitly recommended by their official repositories. For closed-source models, evaluations are conducted via their official API endpoints. We utilize the default generation parameters (\emph{e.g.}, default temperature and top-\(p\) sampling) provided by the respective platforms to accurately reflect their out-of-the-box performance.

\textbf{Evaluation Protocols.} We employ GPT-5 \cite{gpt52025} as the backend for computing the LLM-as-Judge VIEScore. To calculate the text Accuracy Rate (AR), we utilize PPOCRv5 \cite{ppocr2020du} to detect and recognize generated text for comparison against the ground truth. However, while PPOCRv5 excels at standard text, it struggles to accurately detect and recognize artistic or highly stylized text. To ensure evaluation accuracy, we manually correct the AR results for the artistic text and layout-aware generation tasks, which contain a substantial amount of stylized text.

\subsection{Evaluation Result}
\subsubsection{Overall Performance}
\label{sec::task_performance}
Table~\ref{Table::exp_across_all_task} presents quantitative evaluation results across all tasks, from which we draw the following observations.

\textbf{Overall performance remains limited.} The closed-source unified model Nano Banana Pro achieves the highest OCRGenScore of \textbf{77.19}, establishing the current state-of-the-art. Open-source specialized models follow: Flux.2-dev scores 70.19 and LongCat-Image 66.39. However, only Nano Banana Pro and Flux.2-dev exceed 70, and most models fall below 60, far from the theoretical maximum of 100. This reveals substantial deficiencies in current generative models' visual text generation and editing capabilities.

\textbf{Architecture comparison: unified vs. specialized.} The two architectural paradigms exhibit contrasting maturity levels. Among unified models, closed-source Nano Banana Pro demonstrates the architecture's high ceiling, while open-source alternatives lag significantly (BAGEL: 59.11, Janus-4o: 29.58), revealing an around 18-47 point gap. Conversely, open-source specialized models (Flux.2-dev, LongCat-Image) match or exceed closed-source counterparts (GPT Image 1.5: 54.00, Seedream 4.5: 63.35), with top specialized models surpassing most unified alternatives. This suggests unified architectures offer higher potential but require substantial engineering and computational resources currently accessible only to closed-source development. Specialized architectures provide more practical pathways to strong performance, particularly for research applications like fine-tuning.

\textbf{Generation-fidelity trade-off.} We observe that existing models excel at tasks that prioritize generative creativity (T2I, image editing) but struggle with restorative ones (document processing, text removal), particularly document dewarping. This stems from training objectives prioritizing perceptual quality over pixel-level fidelity, causing hallucinations, \emph{i.e.}, plausible but incorrect content such as fabricated text or over-smoothed images. Excellence in generation does not transfer to restoration, highlighting the need for improved fidelity in precision-critical tasks.

\begin{table*}[t]
	\renewcommand{\arraystretch}{1.2}
	\centering
	\caption{Performance comparison on OCRGenBench across different text categories.}
	\resizebox{\linewidth}{!}{
		\begin{tabular}{l c c c c c c c c}
			\toprule
			\multirow{2.2}{*}{\textbf{Model}} & \multicolumn{2}{c}{\textbf{Document}} & \multirow{2.2}{*}{\textbf{Handwriting} $\uparrow$} & \multirow{2.2}{*}{\textbf{Scene Text} $\uparrow$} & \multirow{2.2}{*}{\textbf{Artistic Text} $\uparrow$} & \multicolumn{3}{c}{\textbf{Layout-Rich Text}}\\
			\cmidrule(lr){2-3}\cmidrule(r){7-9}
			& \textbf{Modern Document} $\uparrow$ & \textbf{Historical Document} $\uparrow$ & & & & \textbf{Slide} $\uparrow$ & \textbf{Poster} $\uparrow$ & \textbf{Layout-Aware Text} $\uparrow$\\
			\hline
			\multicolumn{9}{l}{\textbf{Closed-Source Model}}\\
			\hline
			\rowcolor{purple!10}
			\multicolumn{9}{l}{\textcolor{purple}{\emph{Unified Und. \& Gen. Model}}}\\
			Nano Banana Pro \cite{nanobananapro2025} & \textbf{70.87} & \textbf{71.37} & \textbf{78.28} & \textbf{84.67} & \textbf{89.11} & \textbf{92.86} & \underline{80.85} & \textbf{94.43}\\
			\rowcolor{deepred!10}
			\multicolumn{9}{l}{\textcolor{deepred}{\emph{Specialized Generation Model}}}\\
			GPT Image 1.5 \cite{gptimage152025} & 36.31 & 56.47 & 57.16 & 60.37 & 86.26 & \underline{80.67} & 63.21 & \underline{91.73}\\
			Seedream 4.5 \cite{seedream2025seedream} & 48.74 & 61.82 & 60.30 & 70.34 & \underline{89.07} & 77.20 & 74.23 & {74.75}\\
			\hline
			\multicolumn{9}{l}{\textbf{Open-Source Model}}\\
			\hline
			\rowcolor{purple!10}
			\multicolumn{9}{l}{\textcolor{purple}{\emph{Unified Und. \& Gen. Model}}}\\
			BAGEL \cite{bagel2025deng} & 56.33 & 47.63 & 55.47 & 63.83 & 40.25 & 47.79 & 44.91 & 54.15\\
			OmniGen2 \cite{omnigen22025wu} & 45.97 & 46.84 & 40.66 & 63.81 & 47.45 & 42.84 & 42.59 & 57.96\\
			InternVL-U \cite{internvlu2026tian} & 36.83 & 29.96 & 47.58 & 49.94 & 57.94 & 52.56 & 41.60 & 69.88\\
			ILLUME+ \cite{illumeplus2025huang} & 33.87 & 19.48 & 32.55 & 36.05 & 13.40 & 34.90 & 21.57 & 5.48\\
			Janus-4o \cite{janus4o2025chen} & 31.75 & 25.88 & 33.28 & 33.51 & 29.09 & 45.85 & 26.94 & 29.43\\
			\rowcolor{deepred!10}
			\multicolumn{9}{l}{\textcolor{deepred}{\emph{Specialized Generation Model}}}\\
			Qwen-Image \cite{qwenimage2025wu} & 45.22 & 56.37 & 52.12 & 58.59 & 87.26 & 75.63 & 70.40 & 91.68\\
			LongCat-Image \cite{longcat2025meituan} & 54.79 & \underline{63.37} & \underline{70.92} & 72.21 & 85.95 & 79.04 & \textbf{82.48} & 89.86\\
			SD-3.5-Large \cite{sd32024esser} & 30.11 & 22.86 & 33.23 & 30.56 & 47.90 & 43.93 & 36.96 & 11.68\\
			Flux.1-Kontext-dev \cite{fluxkontext2025} & 31.10 & 35.72 & 33.39 & 32.38 & 52.95 & 38.09 & 27.40 & 38.76\\
			Flux.2-dev \cite{flux22025} & \underline{60.17} & 61.36 & 69.75 & \underline{75.50} & 87.04 & 80.17 & 70.68 & 80.52\\
			Flux.2-Klein-9B \cite{flux2klein2026} & 47.85 & 56.55 & 60.73 & 66.89 & 72.56 & 68.01 & 57.04 & 57.83\\
			GLM-Image \cite{glmimage2026} & 41.42 & 62.62 & 57.83 & 49.74 & 68.66 & 66.84 & 64.12 & 73.74\\
			\noalign{\vspace{-1pt}}
			\bottomrule
	\end{tabular}}
	\label{Table::exp_across_text}
\end{table*}

\subsubsection{Performance across Text Categories}
\label{sec::text_performance}
Table~\ref{Table::exp_across_text} presents results across text categories. Nano Banana Pro achieves the highest scores in five of seven categories. Among open-source models, Flux.2-dev and LongCat-Image each lead one category (artistic text and poster) and rank second in two others. Specialized models consistently outperform unified ones, particularly for historical documents, handwriting, and layout-rich text, aligning with task-specific analysis in Sec.~\ref{sec::task_performance}.

Generation difficulty varies significantly across categories. Documents pose the greatest challenge, especially modern documents, where dewarping and deblurring cause substantial performance drops. While performance in historical documents is relatively better, generating a large amount of characters on ancient scripts and locating small text within dense content remains difficult. Handwriting ranks second in difficulty (no model exceeds 80), challenged by high stylistic variability, irregular stroke boundaries, varying intensities in editing tasks, and complications from interleaved printed and handwritten text in removal tasks. Other categories (scene text, layout-rich text) resemble natural scene generation where current models excel, yielding more promising results.
%For CLR text, despite having denser texts, we find that the general performance on slide is better than on poster, 

\begin{table*}[t]
	\renewcommand{\arraystretch}{1.2}
	\centering
	\caption{Performance comparison on OCRGenBench across only T2I generation tasks. Models marked with * support only T2I generation and do not have image editing capabilities. Since slide images contain additional generated text, we refrain from assessing the text accuracy (AR).}
	\resizebox{\textwidth}{!}{
		\begin{tabular}{l c c c c c c c c c c c c c}
			\toprule
			\multirow{2.2}{*}{\textbf{Model}} & \multirow{2.2}{*}{\textbf{OCRGenScore$_\text{T2I}$} $\uparrow$} & \multicolumn{2}{c}{\textbf{Historical Document}} & \multicolumn{2}{c}{\textbf{Handwriting}} & \multicolumn{2}{c}{\textbf{Scene Text}} & \multicolumn{2}{c}{\textbf{Artistic Text}} & \multicolumn{2}{c}{\textbf{Slide}} & \multicolumn{2}{c}{\textbf{Poster}}\\
			\cmidrule(lr){3-4}\cmidrule(lr){5-6}\cmidrule(lr){7-8}\cmidrule(lr){9-10}\cmidrule(lr){11-12}\cmidrule(lr){13-14}
			~ & ~ & \textbf{VIEScore} $\uparrow$ & \textbf{AR} $\uparrow$ & \textbf{VIEScore} $\uparrow$ & \textbf{AR} $\uparrow$ & \textbf{VIEScore} $\uparrow$ & \textbf{AR} $\uparrow$ & \textbf{VIEScore} $\uparrow$ & \textbf{AR} $\uparrow$ & \textbf{VIEScore} $\uparrow$ & \textbf{AR} $\uparrow$ & \textbf{VIEScore} $\uparrow$ & \textbf{AR} $\uparrow$\\
			\hline
			\multicolumn{14}{l}{\textbf{Closed-Source Model}}\\
			\hline
			\rowcolor{purple!10}
			\multicolumn{14}{l}{\textcolor{purple}{\emph{Unified Und. \& Gen. Model}}}\\
			Nano Banana Pro \cite{nanobananapro2025} & \textbf{85.53} & \underline{92.02} & \textbf{61.76} & \underline{91.36} & \textbf{74.35} & \textbf{94.06} & 90.89 & 90.91 & \underline{91.32} & \textbf{94.82} & - & 91.87 & 58.22\\
			\rowcolor{deepred!10}
			\multicolumn{14}{l}{\textcolor{deepred}{\emph{Specialized Generation Model}}}\\
			GPT Image 1.5 \cite{gptimage152025} & {82.63} & \textbf{92.35} & 49.63 & \textbf{94.10} & 67.74 & \underline{93.39} & 83.19 & \underline{92.27} & 77.29 & \underline{94.25} & - & \textbf{93.62} & 59.43\\
			Seedream 4.5 \cite{seedream2025seedream} & \underline{83.83} & 89.60 & 53.38 & 89.48 & 53.92 & 90.81 & 84.58 & 90.98 & \textbf{95.42} & 90.72 & - & \underline{92.50} & \underline{83.87}\\
			\hline
			\multicolumn{14}{l}{\textbf{Open-Source Model}}\\
			\hline
			\rowcolor{purple!10}
			\multicolumn{14}{l}{\textcolor{purple}{\emph{Unified Und. \& Gen. Model}}}\\
			BAGEL \cite{bagel2025deng} & 37.34 & 54.34 & 8.59 & 54.11 & 16.05 & 65.46 & 9.53 & 55.08 & 25.76 & 43.69 & - & 70.40 & 1.40\\
			OmniGen2 \cite{omnigen22025wu} & 45.58 & 73.25 & 18.98 & 50.76 & 24.48 & 69.48 & 32.71 & 58.47 & 37.20 & 48.42 & - & 72.77 & 12.08\\
			InternVL-U \cite{internvlu2026tian} & 58.44 & 58.28 & 10.27 & 56.66 & 40.59 & 69.78 & 80.09 & 83.89 & 46.83 & 76.11 & - & 53.95 & 48.72\\
			ILLUME+ \cite{illumeplus2025huang} & 27.26 & 59.42 & 8.37 & 43.87 & 7.81 & 51.07 & 2.72 & 24.62 & 4.12 & 36.13 & - & 52.72 & 0.14\\
			Janus-4o \cite{janus4o2025chen} & 38.79 & 72.78 & 10.01 & 49.22 & 14.59 & 53.66 & 15.70 & 42.48 & 19.20 & 62.31 & - & 62.30 & 0.87\\
			Show-o2* \cite{showo22025} & 19.44 & 28.71 & 5.01 & 25.83 & 11.26 & 24.79 & 5.47 & 35.14 & 23.67 & 20.07 & - & 33.22 & 0.07\\
			\rowcolor{deepred!10}
			\multicolumn{14}{l}{\textcolor{deepred}{\emph{Specialized Generation Model}}}\\
			Qwen-Image \cite{qwenimage2025wu} & 76.03 & 87.40 & 26.36 & 77.43 & 60.28 & 88.01 & 76.77 & 89.96 & 90.53 & 86.49 & - & 84.32 & 58.30\\
			LongCat-Image \cite{longcat2025meituan} & 78.92 & 82.91 & 42.03 & 78.30 & 51.74 & 87.47 & \textbf{95.99} & 88.55 & 88.05 & 83.82 & - & 88.41 & 75.96\\
			SD-3.5-Large \cite{sd32024esser} & 43.36 & 52.39 & 7.40 & 41.20 & 19.86 & 47.48 & 42.44 & 59.65 & 43.76 & 58.06 & - & 64.30 & 25.70\\
			Flux.1-Kontext-dev \cite{fluxkontext2025} & 31.78 & 37.21 & 9.55 & 36.20 & 21.25 & 40.15 & 29.01 & 39.80 & 33.69 & 39.25 & - & 51.98 & 3.96\\
			Flux.2-dev \cite{flux22025} & 79.28 & 85.59 & 31.41 & 86.94 & 61.81 & 89.19 & 85.70 & \textbf{92.47} & 81.14 & 89.75 & - & 89.97 & 67.62\\
			Flux.2-Klein-9B \cite{flux2klein2026} & 64.38 & 81.89 & 29.94 & 79.47 & 39.99 & 84.13 & 39.24 & 85.91 & 50.27 & 85.32 & - & 83.84 & 27.23\\
			GLM-Image \cite{glmimage2026} & 77.95 & 79.85 & \underline{60.99} & 77.95 & 64.16 & 88.51 & 82.17 & 88.64 & 75.77 & 78.70 & - & 90.49 & 69.48\\
			Z-Image* \cite{zimage2025} & 74.46 & 66.22 & 57.31 & 68.52 & \underline{72.78} & 78.87 & \underline{93.08} & 77.29 & 89.80 & 66.44 & - & 79.48 & 77.30\\
			Ovis-Image* \cite{ovis2025wang} & 69.02 & 52.45 & 52.75 & 55.30 & 54.00 & 76.67 & 85.82 & 77.38 & 92.70 & 57.14 & - & 75.97 & \textbf{90.87}\\
			\noalign{\vspace{-1pt}}
			\bottomrule
	\end{tabular}}
	\label{Table::exp_across_T2I_task}
\end{table*}

\subsubsection{Text-to-Image (T2I) Generation Performance}
Since T2I generation spans across multiple text categories, we break down models' T2I performance across different categories, with results presented in Table~\ref{Table::exp_across_T2I_task}. While Nano Banana Pro achieves the highest OCRGenScore$_\text{T2I}$ of 85.53, specialized models demonstrate concentrated excellence. Closed-source Seedream 4.5 (83.83) and GPT Image 1.5 (82.63) perform exceptionally well, while the open-source Flux.2-dev (79.28) and LongCat-Image (78.92) significantly outperform all unified alternatives. Despite Nano Banana Pro's lead, specialized models currently offer the best performance-to-complexity ratio for T2I text generation.

In terms of visual perceptual quality and semantic fidelity (VIEScore), both unified and specialized models can generally obtain considerable performance. This indicates capable instruction comprehension and visually appealing content generation. However, text accuracy (AR) remains poor across models, revealing insufficient text rendering ability. Historical documents and handwriting prove most challenging due to dense, large-volume text and character complexity (\emph{e.g.}, traditional Chinese). Poster generation also presents difficulties due to the requirement of generating large and small characters with diverse layouts.

% GPT??????
% banana poster ar ???????
% ???????????????????
% unified model?specialized model????????

\begin{table*}[t]
	\renewcommand{\arraystretch}{1.2}
	\centering
	\caption{Performance comparison on OCRGenBench across only text editing tasks. {Doc.} denotes Document. Models marked with * currently support only image editing tasks.}
	\resizebox{\textwidth}{!}{
		\begin{tabular}{l c c c c c c c c c c c c c c c}
			\toprule
			\multirow{3.3}{*}{\textbf{Model}} & \multirow{3.3}{*}{\textbf{OCRGenScore$_\text{Edit}$} $\uparrow$} & \multicolumn{4}{c}{\textbf{Document}} & \multicolumn{2}{c}{\multirow{2.2}{*}{\textbf{Handwriting}}} & \multicolumn{2}{c}{\multirow{2.2}{*}{\textbf{Scene Text}}} & \multicolumn{2}{c}{\multirow{2.2}{*}{\textbf{Artistic Text}}} & \multicolumn{2}{c}{\multirow{2.2}{*}{\textbf{Slide}}} & \multicolumn{2}{c}{\multirow{2.2}{*}{\textbf{Poster}}}\\
			\cmidrule(lr){3-6}
			~ & ~ & \multicolumn{2}{c}{\textbf{Modern Doc.}} & \multicolumn{2}{c}{\textbf{Historical Doc.}} &&&&&&&&&&\\
			\cmidrule(lr){3-4}\cmidrule(lr){5-6}\cmidrule(lr){7-8}\cmidrule(lr){9-10}\cmidrule(lr){11-12}\cmidrule(lr){13-14}\cmidrule(lr){15-16}
			~ & ~ & \textbf{1-LPIPS} $\uparrow$ & \textbf{AR} $\uparrow$ & \textbf{1-LPIPS} $\uparrow$ & \textbf{AR} $\uparrow$ & \textbf{1-LPIPS} $\uparrow$ & \textbf{AR} $\uparrow$ & \textbf{1-LPIPS} $\uparrow$ & \textbf{AR} $\uparrow$ & \textbf{1-LPIPS} $\uparrow$ & \textbf{AR} $\uparrow$ & \textbf{1-LPIPS} $\uparrow$ & \textbf{AR} $\uparrow$ & \textbf{1-LPIPS} $\uparrow$ & \textbf{AR} $\uparrow$\\
			\hline
			\multicolumn{16}{l}{\textbf{Closed-Source Model}}\\
			\hline
			\rowcolor{purple!10}
			\multicolumn{16}{l}{\textcolor{purple}{\emph{Unified Und. \& Gen. Model}}}\\
			Nano Banana Pro \cite{nanobananapro2025} & \underline{80.30} & \underline{88.15} & \textbf{80.23} & 74.66 & \underline{38.87} & 77.42 & \underline{58.23} & \textbf{92.03} & \textbf{73.36} & \textbf{92.15} & 94.00 & \underline{85.92} & \textbf{95.90} & \underline{93.50} & \textbf{79.80}\\
			\rowcolor{deepred!10}
			\multicolumn{16}{l}{\textcolor{deepred}{\emph{Specialized Generation Model}}}\\
			GPT Image 1.5 \cite{gptimage152025} & 52.32 & 55.66 & 28.30 & 47.48 & 8.80 & 54.08 & 31.89 & 59.92 & 45.78 & 67.65 & \underline{97.75} & 66.20 & {67.98} & 55.36 & 44.42\\
			Seedream 4.5 \cite{seedream2025seedream} & 56.74 & 65.14 & 41.82 & 60.95 & 14.71 & 53.77 & 24.68 & 68.67 & 49.51 & 67.15 & \textbf{100.00} & 64.06 & 63.31 & 59.84 & 60.71\\
			\hline
			\multicolumn{16}{l}{\textbf{Open-Source Model}}\\
			\hline
			\rowcolor{purple!10}
			\multicolumn{16}{l}{\textcolor{purple}{\emph{Unified Und. \& Gen. Model}}}\\
			BAGEL \cite{bagel2025deng} & 50.57 & {87.58} & 9.00 & \underline{79.25} & 2.50 & {86.27} & 15.95 & 90.55 & 20.66 & 87.03 & 17.63 & 83.69 & 20.09 & 92.57 & 15.28\\
			OmniGen2 \cite{omnigen22025wu} & 39.86 & 61.76 & 9.32 & 50.00 & 1.88 & 62.57 & 11.60 & 80.30 & 14.22 & 74.28 & 32.01 & 49.88 & 24.66 & 76.14 & 9.39\\
			InternVL-U \cite{internvlu2026tian} & 35.03 & 51.20 & 5.59 & 43.98 & 10.75 & 58.54 & 15.32 & 58.50 & 21.29 & 58.39 & 45.15 & 44.29 & 13.72 & 49.59 & 14.15\\
			ILLUME+ \cite{illumeplus2025huang} & 22.68 & 39.25 & 1.25 & 32.97 & 1.39 & 41.46 & 4.28 & 43.85 & 3.46 & 48.29 & 0.62 & 63.53 & 3.78 & 32.74 & 0.69\\
			Janus-4o \cite{janus4o2025chen} & 26.09 & 43.82 & 2.42 & 32.06 & 1.92 & 44.59 & 3.89 & 47.29 & 6.24 & 52.01 & 27.68 & 56.12 & 2.68 & 43.76 & 0.82\\
			\rowcolor{deepred!10}
			\multicolumn{16}{l}{\textcolor{deepred}{\emph{Specialized Generation Model}}}\\
			Qwen-Image \cite{qwenimage2025wu} & 57.68 & 67.63 & 33.60 & 53.52 & 20.07 & 51.50 & 25.06 & 72.08 & 38.58 & 84.90 & 92.08 & 72.98 & 56.57 & 77.87 & 61.13\\
			LongCat-Image \cite{longcat2025meituan} & 70.97 & 84.35 & 29.83 & 76.73 & 32.06 & \underline{86.55} & {52.73} & 87.56 & 54.12 & 83.62 & 91.95 & 83.81 & 64.72 & 92.90 & 72.66\\
			SD-3.5-Large \cite{sd32024esser} & 27.61 & 46.37 & 4.45 & 37.15 & 0.38 & 43.19 & 5.87 & 42.86 & 5.56 & 64.87 & 18.46 & 59.60 & 0.00 & 51.50 & 6.33\\
			Flux.1-Kontext-dev \cite{fluxkontext2025} & 35.48 & 51.49 & 6.82 & 45.40 & 1.33 & 54.66 & 11.57 & 53.77 & 11.27 & 62.81 & 70.12 & 57.29 & 16.58 & 50.25 & 3.39\\
			Flux.2-dev \cite{flux22025} & 64.14 & 83.53 & 34.86 & 74.61 & 13.08 & 80.93 & 31.32 & 87.57 & 48.38 & {87.58} & 89.77 & {84.51} & 56.67 & 91.28 & 33.86\\
			Flux.2-Klein-9B \cite{flux2klein2026} & 56.00 & 75.19 & 22.29 & 68.29 & 5.67 & 85.01 & 28.46 & 84.66 & 34.58 & 84.39 & 76.90 & 71.35 & 30.07 & 91.18 & 25.93\\
			GLM-Image \cite{glmimage2026} & 48.32 & 76.15 & 16.35 & 67.51 & 1.05 & 59.91 & 15.77 & 72.69 & 24.61 & 82.57 & 53.34 & 75.98 & 33.96 & 79.75 & 16.77\\
			FireRed-Image-Edit-v1.1* \cite{firered2026} & \textbf{81.27} & \textbf{89.75} & \underline{60.08} & \textbf{80.46} & \textbf{44.70} & \textbf{90.68} & \textbf{76.18} & \underline{91.97} & \underline{66.57} & \underline{91.18} & 93.26 & \textbf{88.62} & \underline{88.66} & \textbf{96.19} & \underline{79.47}\\
			\noalign{\vspace{-1pt}}
			\bottomrule
	\end{tabular}}
	\label{Table::exp_across_editing_task}
\end{table*}

\subsubsection{Text Editing Performance}
Similar to T2I generation, Table~\ref{Table::exp_across_editing_task} breaks down text editing performance across categories. Only FireRed-Image-Edit-v1.1 and Nano Banana Pro achieve OCRGenScore$_\text{Edit}$ above 80, with FireRed-Image-Edit-v1.1 performing best overall. Among open-source models, LongCat-Image and Flux.2-dev perform best. Notably, most specialized models surpass a score of 50, whereas Nano Banana Pro and BAGEL are the only unified models to do so. This highlights a distinct weakness of current open-source unified models in visual text modification.

AR metrics indicate that scene text and artistic text are relatively easier to edit with higher success rates. Other categories pose greater challenges, particularly text-dense content (modern/historical documents, slides) where extensive text complicates target localization, yielding incorrect or unmodified results. A more detailed analysis is presented in Sec.~\ref{sec::discuss}, Finding 1. Additionally, for handwriting with extreme aspect ratios (width:height $>$ 8:1), many models (Nano Banana Pro, GPT Image 1.5, GLM-Image, Qwen-Image) regenerate new images rather than editing originals, indicating room for improvement for handwriting images.

\begin{table*}[t]
	\renewcommand{\arraystretch}{1.2}
	\caption{Performance comparison on OCRGenBench across all tasks in Chinese and English.}
	\begin{minipage}{\textwidth}
		\centering
		\resizebox{\linewidth}{!}{
			\begin{tabular}{l c c c c c c c c c c c c c c c c c c}
				\toprule
				\multirow{3.3}{*}{\textbf{Model}} & \multicolumn{2}{c}{\multirow{2.2}{*}{\textbf{OCRGenScore} $\uparrow$}} & \multicolumn{4}{c}{\textbf{T2I}} & \multicolumn{4}{c}{\textbf{Editing}} & \multicolumn{2}{c}{\textbf{Doc. Dewarping}} & \multicolumn{2}{c}{\textbf{Doc. Deshadowing}} & \multicolumn{2}{c}{\textbf{Doc. Deblurring}} & \multicolumn{2}{c}{\textbf{Appear. Enhance.}}\\
				\cmidrule(lr){4-7}\cmidrule(lr){8-11}\cmidrule(lr){12-13}\cmidrule(lr){14-15}\cmidrule(lr){16-17}\cmidrule(lr){18-19}
				~ & ~ & ~ & \multicolumn{2}{c}{\textbf{VIEScore} $\uparrow$} & \multicolumn{2}{c}{\textbf{AR} $\uparrow$} & \multicolumn{2}{c}{\textbf{1-LPIPS} $\uparrow$} & \multicolumn{2}{c}{\textbf{AR} $\uparrow$} & \multicolumn{2}{c}{\textbf{DD} $\uparrow$} & \multicolumn{2}{c}{\textbf{MSSSIM} $\uparrow$} & \multicolumn{2}{c}{\textbf{MSSSIM} $\uparrow$} & \multicolumn{2}{c}{\textbf{MSSSIM} $\uparrow$}\\
				\cmidrule(lr){2-3}\cmidrule(lr){4-5}\cmidrule(lr){6-7}\cmidrule(lr){8-9}\cmidrule(lr){10-11}\cmidrule(lr){12-13}\cmidrule(lr){14-15}\cmidrule(lr){16-17}\cmidrule(lr){18-19}
				~ & \textbf{zh} & \textbf{en} & \textbf{zh} & \textbf{en} & \textbf{zh} & \textbf{en} & \textbf{zh} & \textbf{en} & \textbf{zh} & \textbf{en} & \textbf{zh} & \textbf{en} & \textbf{zh} & \textbf{en} & \textbf{zh} & \textbf{en} & \textbf{zh} & \textbf{en}\\
				\hline
				\multicolumn{19}{l}{\textbf{Closed-Source Model}}\\
				\hline
				\rowcolor{purple!10}
				\multicolumn{19}{l}{\textcolor{purple}{\emph{Unified Und. \& Gen. Model}}}\\
				Nano Banana Pro \cite{nanobananapro2025} & \textbf{76.02} & \textbf{78.75} & \underline{91.61} & \underline{92.87} & \textbf{82.27} & \underline{71.74} & 83.43 & \underline{87.00} & \textbf{67.96} & \textbf{74.96} & \textbf{43.99} & \underline{41.53} & \textbf{93.82} & \textbf{81.99} & \textbf{60.26} & \textbf{62.92} & \textbf{79.25} & \textbf{81.19}\\
				\rowcolor{deepred!10}
				\multicolumn{19}{l}{\textcolor{deepred}{\emph{Specialized Generation Model}}}\\
				GPT Image 1.5 \cite{gptimage152025} & 53.04 & 53.95 & \textbf{93.50} & \textbf{93.31} & \underline{69.24} & 68.21 & 55.35 & 59.36 & 34.08 & 51.37 & 31.69 & 28.13 & 34.34 & 36.99 & 33.60 & 29.98 & 42.50 & 41.26\\
				Seedream 4.5 \cite{seedream2025seedream} & 63.55 & 62.73 & 91.09 & 89.82 & 66.89 & \textbf{75.22} & 61.87 & 61.62 & 40.88 & 49.51 & 21.45 & 25.45 & \underline{83.87} & 45.05 & 49.51 & \underline{62.90} & 54.42 & 52.36\\
				\hline
				\multicolumn{19}{l}{\textbf{{Open-Source Model}}}\\
				\hline
				\rowcolor{purple!10}
				\multicolumn{19}{l}{\textcolor{purple}{\emph{Unified Und. \& Gen. Model}}}\\
				BAGEL \cite{bagel2025deng} & 55.87 & 60.96 & 53.09 & 61.02 & 2.32 & 27.37 & \textbf{86.09} & \textbf{87.96} & 8.97 & 21.16 & 19.25 & 21.84 & 74.19 & \underline{77.17} & 59.08 & 61.57 & \underline{74.70} & \underline{77.55}\\
				OmniGen2 \cite{omnigen22025wu} & 47.82 & 62.26 & 51.27 & 68.53 & 0.87 & 51.47 & 63.12 & 68.54 & 4.99 & 22.66 & 19.86 & 22.11 & 68.38 & 62.20 & 46.71 & 39.44 & 58.35 & 70.76\\
				InternVL-U \cite{internvlu2026tian} & 43.88 & 45.45 & 70.18 & 62.36 & 44.30 & 44.04 & 50.93 & 56.82 & 9.34 & 25.08 & 23.52 & 32.13 & 64.68 & 41.83 & 33.54 & 26.97 & 49.36 & 44.75\\
				ILLUME+ \cite{illumeplus2025huang} & 29.17 & 31.02 & 32.99 & 53.17 & 0.50 & 10.34 & 42.34 & 43.09 & 1.69 & 3.67 & 22.80 & 30.16 & 44.82 & 41.75 & 42.50 & 38.54 & 35.76 & 40.34\\
				Janus-4o \cite{janus4o2025chen} & 28.50 & 33.54 & 43.85 & 63.19 & 0.19 & 26.86 & 44.02 & 47.04 & 1.17 & 10.43 & 25.94 & 24.18 & 31.45 & 35.70 & 39.39 & 33.95 & 42.06 & 37.10\\
				\rowcolor{deepred!10}
				\multicolumn{19}{l}{\textcolor{deepred}{\emph{Specialized Generation Model}}}\\
				Qwen-Image \cite{qwenimage2025wu} & 56.44 & 55.14 & 88.11 & 80.76 & 68.43 & 62.04 & 60.69 & 70.61 & 33.06 & 49.34 & 19.27 & 29.06 & 60.62 & 44.77 & 42.47 & 33.67 & 66.59 & 57.34\\
				LongCat-Image \cite{longcat2025meituan} & 65.97 & 65.76 & 87.39 & 80.70 & 68.92 & 66.12 & \underline{85.83} & 85.29 & \underline{52.70} & \underline{56.37} & 28.00 & 28.07 & 79.45 & 68.45 & 59.62 & 60.66 & 56.43 & 56.78\\
				SD-3.5-Large \cite{sd32024esser} & 26.74 & 35.32 & 22.55 & 78.96 & 0.60 & 54.00 & 45.20 & 49.65 & 1.04 & 10.50 & 31.62 & 30.57 & 22.71 & 32.24 & 37.63 & 25.66 & 29.06 & 35.23\\
				Flux.1-Kontext-dev \cite{fluxkontext2025} & 28.52 & 43.59 & 5.85 & 72.89 & 0.22 & 42.82 & 53.45 & 54.07 & 2.90 & 27.37 & 19.78 & 27.31 & 26.00 & 32.54 & 34.56 & 25.53 & 41.12 & 40.71\\
				Flux.2-dev \cite{flux22025} & \underline{69.30} & \underline{70.74} & 89.60 & 88.17 & 58.42 & 74.19 & 81.81 & 86.04 & 28.48 & 54.64 & \underline{37.82} & \textbf{43.81} & 81.94 & 60.98 & \underline{61.19} & 59.35 & 73.15 & 70.82\\
				Flux.2-Klein-9B \cite{flux2klein2026} & 56.53 & 61.95 & 80.47 & 85.18 & 11.39 & 67.43 & 78.95 & 83.40 & 12.37 & 49.82 & 23.16 & 25.28 & 83.86 & 45.04 & 51.34 & 51.37 & 62.81 & 51.78\\
				GLM-Image \cite{glmimage2026} & 46.72 & 47.17 & 85.98 & 81.11 & 63.47 & 75.16 & 70.68 & 71.25 & 13.82 & 29.26 & 21.44 & 27.36 & 25.73 & 51.86 & 44.02 & 37.08 & 40.70 & 60.51\\
				\noalign{\vspace{-1pt}}
				\bottomrule
		\end{tabular}}
	\end{minipage}
	\vfill
	\vspace{3pt}
	\begin{minipage}{\textwidth}
		\centering
		\resizebox{\linewidth}{!}{
			\begin{tabular}{l c c c c c c c c c c c c c c c c c c c c}
				\toprule
				\multirow{4.4}{*}{\textbf{Model}} & \multicolumn{2}{c}{\multirow{3.3}{*}{\textbf{OCRGenScore} $\uparrow$}} & \multicolumn{4}{c}{\textbf{Text Removal}} & \multicolumn{6}{c}{\textbf{Style Transfer}} & \multicolumn{2}{c}{\multirow{2.2}{*}{\textbf{Hist. Doc. Rest.}}} & \multicolumn{2}{c}{\multirow{2.2}{*}{\textbf{Scene Text Super Resolution}}} & \multicolumn{4}{c}{\multirow{2.2}{*}{\textbf{Layout-Aware Text Gen.}}}\\
				\cmidrule(lr){4-7}\cmidrule(lr){8-13}
				~ & ~ & ~ & \multicolumn{2}{c}{\textbf{Handwriting}} & \multicolumn{2}{c}{\textbf{Scene Text}} & \multicolumn{4}{c}{\textbf{Artistic Text}} & \multicolumn{2}{c}{\textbf{Hist. Doc.}} &&&&&&&&\\
				\cmidrule(lr){4-5}\cmidrule(lr){6-7}\cmidrule(lr){8-11}\cmidrule(lr){12-13}\cmidrule(lr){14-15}\cmidrule(lr){16-17}\cmidrule(lr){18-19}\cmidrule(lr){20-21}
				~ & ~ & ~ & \multicolumn{2}{c}{\textbf{MSSSIM} $\uparrow$} & \multicolumn{2}{c}{\textbf{MSSSIM} $\uparrow$} & \multicolumn{2}{c}{\textbf{VIEScore} $\uparrow$} & \multicolumn{2}{c}{\textbf{AR} $\uparrow$} & \multicolumn{2}{c}{\textbf{VIEScore} $\uparrow$} & \multicolumn{2}{c}{\textbf{1-LPIPS} $\uparrow$} & \multicolumn{2}{c}{\textbf{MSSSIM} $\uparrow$} & \multicolumn{2}{c}{\textbf{VIEScore} $\uparrow$} & \multicolumn{2}{c}{\textbf{AR} $\uparrow$}\\
				\cmidrule(lr){2-3}\cmidrule(lr){4-5}\cmidrule(lr){6-7}\cmidrule(lr){8-9}\cmidrule(lr){10-11}\cmidrule(lr){12-13}\cmidrule(lr){14-15}\cmidrule(lr){16-17}\cmidrule(lr){18-19}\cmidrule(lr){20-21}
				~ & \textbf{zh} & \textbf{en} & \textbf{zh} & \textbf{en} & \textbf{zh} & \textbf{en} & \textbf{zh} & \textbf{en} & \textbf{zh} & \textbf{en} & \textbf{zh} & \textbf{en} & \textbf{zh} & \textbf{en} & \textbf{zh} & \textbf{en} & \textbf{zh} & \textbf{en} & \textbf{zh} & \textbf{en}\\
				\hline
				\multicolumn{21}{l}{\textbf{Closed-Source Model}}\\
				\hline
				\rowcolor{purple!10}
				\multicolumn{21}{l}{\textcolor{purple}{\emph{Unified Und. \& Gen. Model}}}\\
				Nano Banana Pro \cite{nanobananapro2025} & \textbf{76.02} & \textbf{78.75} & \textbf{84.16} & - & \textbf{93.53} & \underline{89.80} & 64.21 & \underline{92.33} & 79.23 & 98.78 & 77.66 & - & \underline{74.15} & - & 52.98 & 76.37 & \textbf{88.02} & \textbf{89.63} & \textbf{100.00} & \textbf{100.00}\\
				\rowcolor{deepred!10}
				\multicolumn{21}{l}{\textcolor{deepred}{\emph{Specialized Generation Model}}}\\
				GPT Image 1.5 \cite{gptimage152025} & 53.04 & 53.95 & 47.59 & - & 50.75 & 52.00 & \textbf{81.63} & \textbf{92.68} & 90.11 & 98.33 & \textbf{80.57} & - & 46.18 & - & 19.68 & 26.79 & \underline{86.05} & {85.36} & 95.30 & \textbf{100.00}\\
				Seedream 4.5 \cite{seedream2025seedream} & 63.55 & 62.73 & 69.97 & - & 88.99 & 78.87 & \underline{76.60} & 86.45 & \textbf{99.23} & \underline{99.44} & \underline{78.07} & - & 59.88 & - & 29.30 & 54.73 & 55.26 & 53.81 & 89.47 & \textbf{100.00}\\
				\hline
				\multicolumn{21}{l}{\textbf{Open-Source Model}}\\
				\hline
				\rowcolor{purple!10}
				\multicolumn{21}{l}{\textcolor{purple}{\emph{Unified Und. \& Gen. Model}}}\\
				BAGEL \cite{bagel2025deng} & 55.87 & 60.96 & \underline{80.23} & - & 87.20 & 72.35 & 12.74 & 53.65 & 1.30 & 44.29 & 44.49 & - & 73.71 & - & \textbf{88.05} & \textbf{87.28} & 82.51 & 68.78 & 3.74 & 59.45\\
				OmniGen2 \cite{omnigen22025wu} & 47.82 & 62.26 & 47.28 & - & 83.19 & 81.03 & 5.66 & 64.68 & 6.52 & 88.65 & 64.66 & - & 50.64 & - & \underline{82.67} & \underline{84.83} & 67.37 & 77.30 & 0.00 & 82.53\\
				InternVL-U \cite{internvlu2026tian} & 43.88 & 45.45 & 57.19 & - & 53.47 & 68.80 & 44.41 & 62.41 & 57.21 & 62.77 & 20.75 & - & 37.45 & - & 19.48 & 16.00 & 62.28 & 46.40 & 83.30 & 88.11\\
				ILLUME+ \cite{illumeplus2025huang} & 29.17 & 31.02 & 48.93 & - & 71.65 & 65.19 & 0.00 & 1.00 & 0.00 & 4.52 & 1.00 & - & 25.84 & - & 16.19 & 27.56 & 3.68 & 11.27 & 0.00 & 6.27\\
				Janus-4o \cite{janus4o2025chen} & 28.50 & 33.54 & 43.69 & - & 37.13 & 34.41 & 0.00 & 50.00 & 0.00 & 16.34 & 16.53 & - & 28.60 & - & 16.21 & 21.88 & 32.86 & 49.41 & 0.00 & 32.98\\
				\rowcolor{deepred!10}
				\multicolumn{21}{l}{\textcolor{deepred}{\emph{Specialized Generation Model}}}\\
				Qwen-Image \cite{qwenimage2025wu} & 56.44 & 55.14 & 49.22 & - & 38.52 & 50.60 & 68.27 & 72.99 & \underline{96.01} & 94.92 & 72.33 & - & 59.48 & - & 27.24 & 28.07 & 82.44 & \underline{86.24} & \underline{99.00} & \underline{98.87}\\
				LongCat-Image \cite{longcat2025meituan} & 65.97 & 65.76 & 78.11 & - & 90.32 & 89.20 & 58.02 & 80.42 & 91.30 & 97.25 & 69.74 & - & 66.86 & - & 18.05 & 30.69 & 82.47 & 84.57 & 95.37 & 96.84\\
				SD-3.5-Large \cite{sd32024esser} & 26.74 & 35.32 & 44.64 & - & 25.62 & 26.60 & 46.03 & 67.24 & 3.91 & 84.10 & 0.00 & - & 42.79 & - & 18.16 & 16.59 & 17.93 & 19.05 & 0.00 & 9.21\\
				Flux.1-Kontext-dev \cite{fluxkontext2025} & 28.52 & 43.59 & 38.34 & - & 27.83 & 35.91 & 37.85 & 79.47 & 8.39 & 96.78 & 52.90 & - & 43.24 & - & 21.34 & 23.64 & 22.61 & 72.88 & 0.00 & 54.32\\
				Flux.2-dev \cite{flux22025} & \underline{69.30} & \underline{70.74} & 78.75 & - & 88.21 & \textbf{91.10} & 73.98 & 85.94 & 82.64 & \textbf{100.00} & 68.48 & - & \textbf{74.62} & - & 53.10 & 52.45 & 75.87 & 77.13 & 76.57 & 91.70\\
				Flux.2-Klein-9B \cite{flux2klein2026} & 56.53 & 61.95 & 65.71 & - & \underline{92.63} & 88.55 & 67.54 & 89.41 & 18.79 & \textbf{100.00} & 67.48 & - & 65.81 & - & 44.49 & 48.43 & 75.16 & 73.04 & 10.49 & 69.77\\
				GLM-Image \cite{glmimage2026} & 46.72 & 47.17 & 64.61 & - & 18.00 & 14.42 & 54.20 & 65.35 & 44.39 & 59.40 & 77.81 & - & 67.98 & - & 12.53 & 11.07 & 78.31 & 83.08 & 58.31 & 74.23\\
				\noalign{\vspace{-1pt}}
				\bottomrule
		\end{tabular}}
	\end{minipage}
	\label{Table::exp_across_all_task_language}
\end{table*}

\section{Discussion}
\label{sec::discuss}
%From the above evaluation across multiple OCR generative tasks, we draw the following findings.
In addition to quantitative results, we perform visualizations to more intuitively observe models' performance. Based on these results, we conduct in-depth analyses to identify the key limitations of current generative models and explore potential directions for improvement. These findings are detailed as follows.

% location???????ppt?, ????????
\begin{tcolorbox}[colback=white, colframe=black, left=1pt, right=1pt, top=1pt, bottom=1pt]
	\textbf{Finding 1.} Models lack sufficient text perception and localization abilities for accurate text editing.
\end{tcolorbox}

\begin{figure*}[t]
	\centering
	\includegraphics[width=\textwidth]{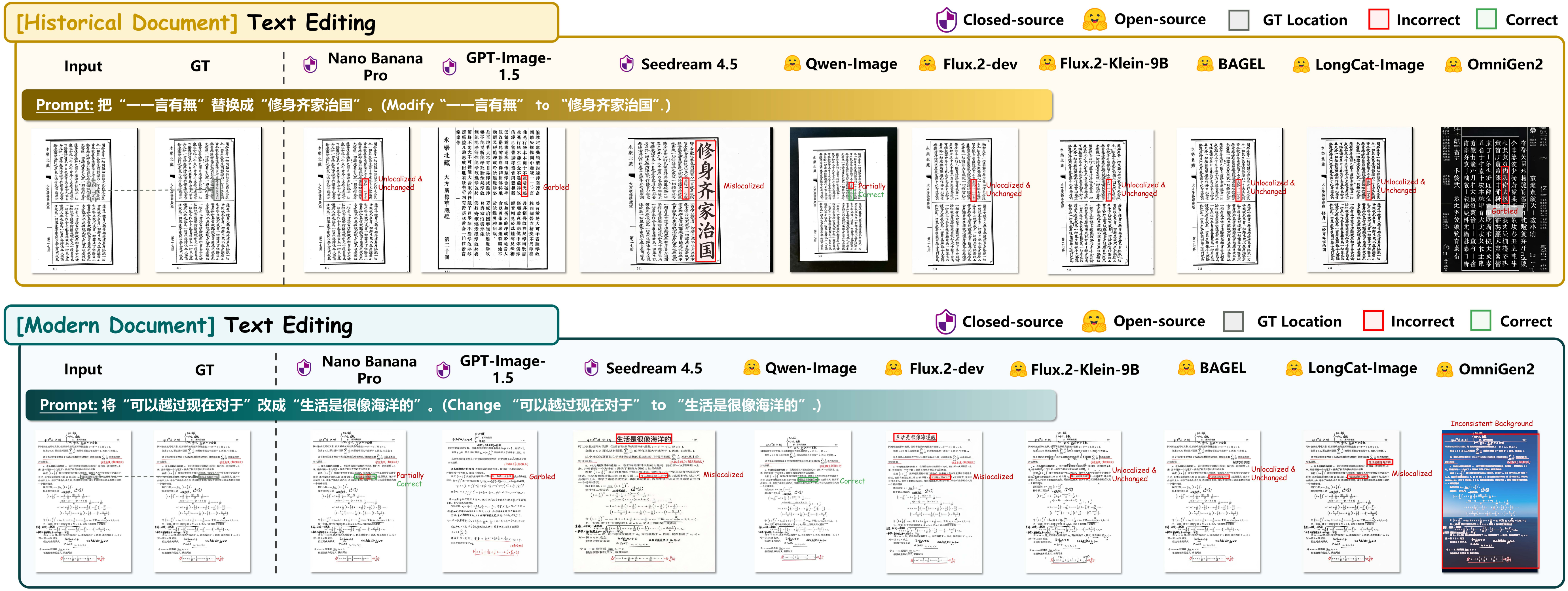}
	\caption{Visualization examples where models fail to locate the target text, leading to incorrect editing.}
	\label{Fig::visualize1}
\end{figure*}

In Tables~\ref{Table::exp_across_all_task} and \ref{Table::exp_across_editing_task}, current models demonstrate poor AR in editing tasks, with the best overall AR only reaching 71.46\% and most category-specific scores below 50\%. This stems from insufficient text localization abilities. When processing dense, extensive text, models struggle to locate target regions, resulting in mislocalized edits or unchanged content.  This is evident in Fig.~\ref{Fig::visualize1}, where most models struggle to locate a single line of text among an extensive page like modern/historical documents, causing incorrect edits. This issue also appears in scene text and naturally embedded text. As text localization is fundamental to editing, this deficiency warrants future research attention.

% ??ok?????????text editing????????????????
\begin{tcolorbox}[colback=white, colframe=black, left=1pt, right=1pt, top=1pt, bottom=1pt]
	\textbf{Finding 2.} Existing models may fail to maintain the consistency of non-target elements.
\end{tcolorbox}

\begin{figure*}[t]
	\centering
	\includegraphics[width=\textwidth]{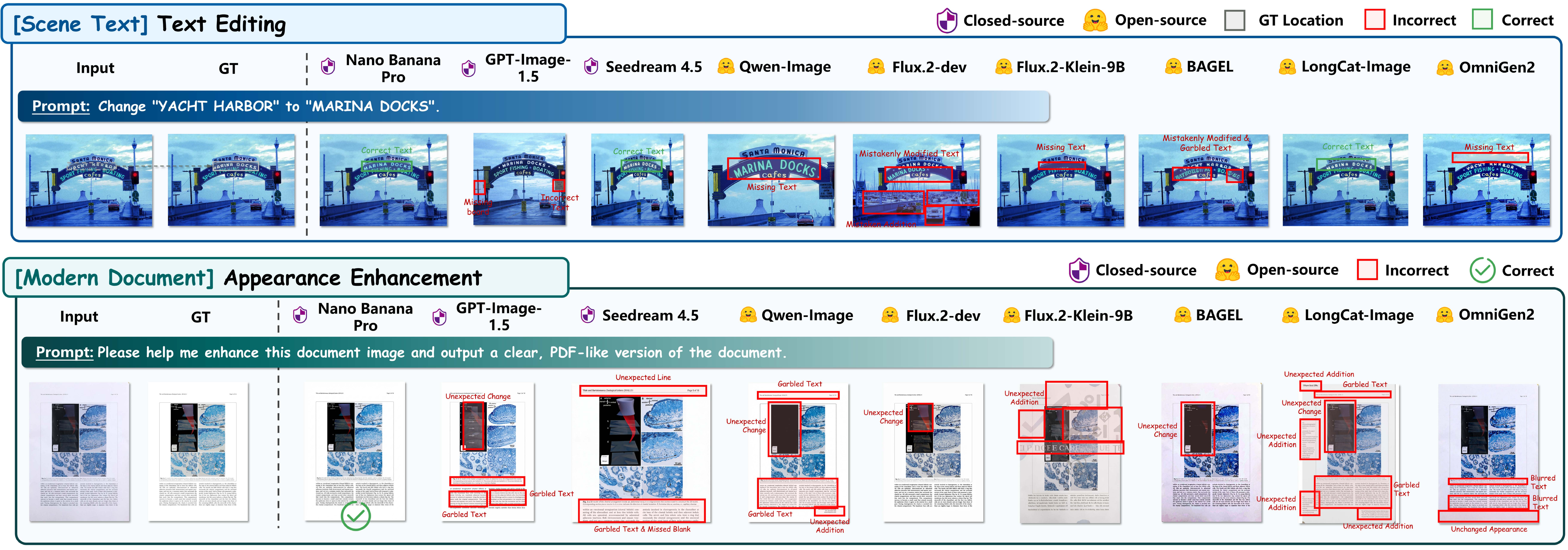}
	\caption{Visualization examples where models fail to preserve other content in the original images, such as visual objects or non-target text.}
	\label{Fig::visualize2}
\end{figure*}

When modifying specific text regions, current models often fail to preserve unchanged elements, including non-target text and visual objects. Fig.~\ref{Fig::visualize2} shows that while models successfully manipulate target text, they inadvertently alter surrounding text (removing or adding characters) and background details. This issue is prevalent in artistic text and poster editing, contributing to unsatisfactory MSSSIM and LPIPS scores in Table~\ref{Table::exp_across_all_task}. This consistency problem requires architectures with explicit spatial control, such as region-aware RL rewards and preservation-focused losses, to localize editing operations to target areas.

\begin{figure*}[t]
	\centering
	\includegraphics[width=\textwidth]{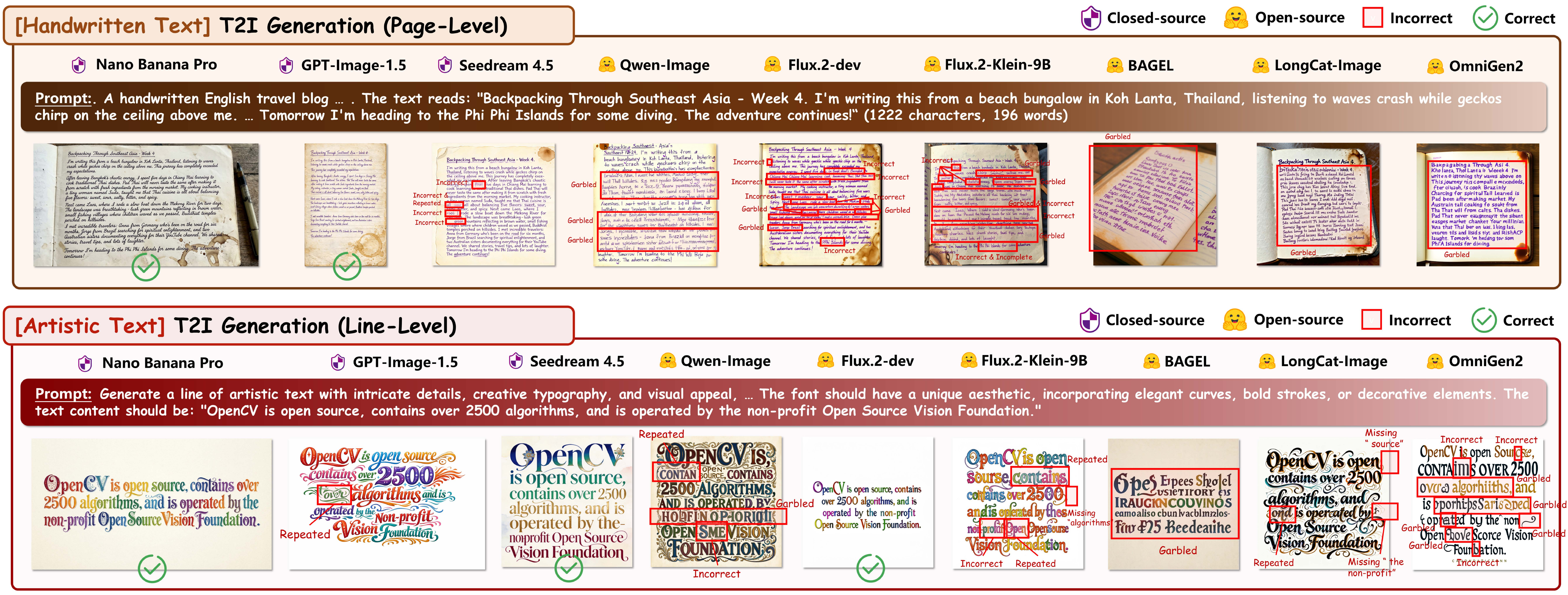}
	\caption{Visualization examples of T2I generation with dense, long text and complex font.}
	\label{Fig::visualize3}
\end{figure*}

\begin{figure*}[t]
	\centering
	\includegraphics[width=\textwidth]{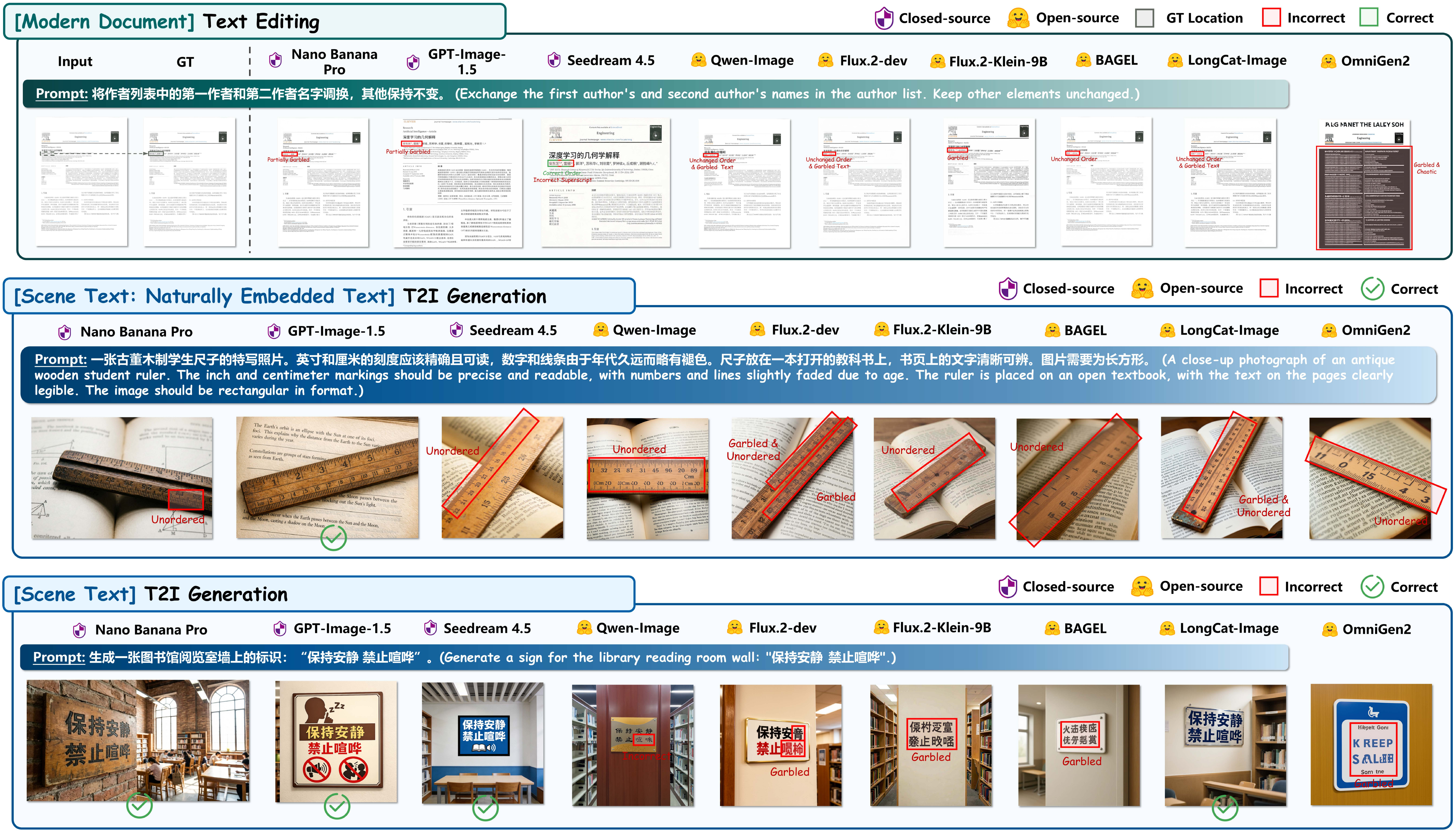}
	\caption{Visualization examples of inferior understanding and inability in generating Chinese text.}
	\label{Fig::visualize4}
\end{figure*}

% instruction-following?Banana??Poster T2I?OmniGen2?,GPT?????????????????hallucination?Flux1?Showo2???illumeplus???????????
\begin{tcolorbox}[colback=white, colframe=black, left=1pt, right=1pt, top=1pt, bottom=1pt]
	\textbf{Finding 3.} Models exhibit unstable instruction-following and sometimes produce hallucinations.
\end{tcolorbox}

Models can fail to follow or understand user instructions, producing unstable outputs. In Table~\ref{Table::exp_across_T2I_task}, despite Nano Banana Pro's strong AR performance, it shows quite worse AR (58.22\%) for poster T2I generation. This is because when the prompt requires generating lowercase English letters, it generates totally uppercase instead, which is an issue also affecting GPT-Image-1.5 and Qwen-Image. In document dewarping, we give the prompt: ``Please help me perform geometric rectification on the document image to restore a flat document that fills the image area.''. Despite this clear directive, most models fail to dewarp in over 40\% of cases, and even second-best Flux.2-dev often produces flat images without filling the entire area. In image-conditioned tasks, GPT-Image-1.5 frequently fails to preserve original aspect ratios, resulting in its underperformance.

Hallucination presents a more severe problem. Models like Flux.1-Kontext-dev, Show-o2, and Illume+ generate irrelevant content (\emph{e.g.}, human figures) when prompted for textual content, while StableDiffusion can produce chaotic pixels indicating collapse. Earlier models (before August 2025) are more prone to hallucinations, likely due to insufficient visual text synthesis data during training. Recently published models incorporate more specific text generation and editing data, effectively reducing such hallucinations.

% ???????????????text??????????????????ppt??????????
\begin{tcolorbox}[colback=white, colframe=black, left=1pt, right=1pt, top=1pt, bottom=1pt]
	\textbf{Finding 4.} Dense and long text synthesis and complex font generation remain unaddressed.
\end{tcolorbox}

In Table~\ref{Table::exp_across_T2I_task}, current models exhibit lower AR under historical documents and handwriting categories. This stems from the large volume and density of required text, which increases error rates. As shown in the top panel of Fig.~\ref{Fig::visualize3}, generating very long handwriting text severely challenges existing models, producing incorrect and garbled text. Similar issues occur in historical document generation. In document-related tasks, high text density causes blurred or garbled output, as seen in document enhancement (Fig.~\ref{Fig::visualize2}) and dewarping. Furthermore, while current models excel at generating line-level text, they struggle with complex fonts even for simple line-level English text, as demonstrated in the bottom panel of Fig.~\ref{Fig::visualize3}. Despite claims of handling extensive, complex text with high accuracy from many models, OCRGenBench evaluation reveals substantial room for improvement.

% understanding???????????
\begin{tcolorbox}[colback=white, colframe=black, left=1pt, right=1pt, top=1pt, bottom=1pt]
	\textbf{Finding 5.} The understanding skills of current generative models demand improvement.
\end{tcolorbox}

Visual text synthesis requires not only content accuracy and visual aesthetics, but also the understanding of world knowledge and common sense logic. However, existing models show poor comprehension skills. As illustrated in the top panel of Fig.~\ref{Fig::visualize4}, when asked to exchange the first and second authors' names, models typically fail to grasp the concepts of ``author name'' and ``name order'', failing to exchange the author names. The middle panel reveals that models lack basic knowledge of ruler numbering and keyboard layouts, producing garbled and chaotic results. They also fail to interpret physical laws required for restorative tasks like document dewarping and super-resolution, as evidenced by the 40+\% failure probability in \emph{Finding 3}. These results underscore the necessity of enhancing models' understanding abilities to unlock more generalizable OCR generation. Nano Banana Pro's superior performance suggests that unified models, which jointly train understanding and generation, offer a promising direction by enabling knowledge transfer between tasks.

\begin{tcolorbox}[colback=white, colframe=black, left=1pt, right=1pt, top=1pt, bottom=1pt]
	\textbf{Finding 6.} Existing models typically fall short in specialized vertical-domain tasks.
\end{tcolorbox}

For highly specialized domain-specific tasks, such as historical document restoration, scene text super-resolution, and low-level document processing (\emph{e.g.}, dewarping), they show generally lower performance than T2I generation or text editing (Table~\ref{Table::exp_across_all_task}). Since existing models primarily focus on these creativity-based tasks, they typically fall short in specialized OCR tasks without domain-specific proficiency. However, incorporating these tasks can broaden the general intelligence of generative models, elevating their generalizability to real-world visual text processing.

% Klein???????BAGEL???Omnigen2?????????
\begin{tcolorbox}[colback=white, colframe=black, left=1pt, right=1pt, top=1pt, bottom=1pt]
	\textbf{Finding 7.} Current models perform well on English samples but fall short in Chinese samples.
\end{tcolorbox}

Table~\ref{Table::exp_across_all_task_language} presents performance across Chinese and English samples. The overall OCRGenScore and most tasks show better English performance. Models like Flux.2-Klein-9B, BAGEL, OmniGen2, and Janus-4o completely fail at Chinese text generation, producing garbled texts (Fig.~\ref{Fig::visualize4}, the bottom panel) and resulting in extremely low AR in T2I generation and text editing. This reflects a pervasive English prioritization in current models, due to large-scale English training data, smaller vocabulary size, and convenient rule-based tokenization. In contrast, Chinese remains poorly supported. Adding multilingual support for Chinese, Arabic, Persian, and other languages is essential for universally applicable, linguistically diverse generative models.

\begin{tcolorbox}[colback=white, colframe=black, left=1pt, right=1pt, top=1pt, bottom=1pt]
	\textbf{Finding 8.} Current models lack fine-grained encoding and decoding capabilities for generating small-scale text.
\end{tcolorbox}

As shown in Table~\ref{Table::exp_across_text} and Sec.~\ref{sec::text_performance}, model performance on modern and historical documents is significantly lower than other types. Beyond the challenges of high-density and extensive text, a key limitation is that existing models cannot encode/decode small-scale characters due to their coarse visual representation mechanisms. As shown in Fig.~\ref{Fig::visualize2}, ~\ref{Fig::visualize3}, and \ref{Fig::visualize4}, models can produce garbled text that looks visually plausible but semantically meaningless. While many models use VAE encoders and decoders, the limited latent dimension size prevents encoding sufficient pixel details, prohibiting faithful generation of strokes and character structures. Hence, enhancing visual encoding and decoding, or exploring novel visual representation mechanisms, represents a promising direction for future research.

\section{Limitation}
While comprehensive and diverse in design, OCRGenBench has several limitations. First, the benchmark's current scale of 1,060 samples may not exhaustively capture the long-tail distribution of rare text styles or extreme corner cases. Still, this scale is adequate for systematic evaluation, and it is comparable to existing benchmarks like TextEditBench (1,169 images) \cite{texteditbench2025gui}, ImgEditBench (734 images) \cite{imgedit2025ye}, and CVTG-2K (2,000 prompts) \cite{cvtg2k2025}. Second, our Accuracy Rate (AR) metric relies on PP-OCRv5 for text recognition; consequently, inherent recognition errors or algorithmic biases, particularly for artistic texts, could inadvertently affect evaluation accuracy. Although we perform manual correction to mitigate this issue (Sec.~\ref{sec::impl}), the dependency on OCR engines remains a constraint. Third, OCRGenBench currently covers only English and Chinese. Expanding the dataset to encompass a broader range of diverse scripts (\emph{e.g.}, Arabic, Korean, or Hindi) can further enhance its comprehensiveness. Finally, current computational resource constraints prevent the evaluation of exceptionally large models (\(\ge\) 60B parameters), which we leave to future work. Despite these limitations, we believe OCRGenBench establishes a solid foundation for systematically evaluating OCR generative capabilities and will continue to evolve with community feedback.

\section{Conclusion}
In this paper, we propose OCRGenBench, a novel benchmark tailored to evaluate image generation models' visual text synthesis abilities. OCRGenBench is the first to unify text-centric T2I generation, text editing, and OCR I2I translation tasks for holistic assessment of models' visual text synthesis skills, \emph{i.e.}, OCR generative capabilities. The benchmark encompasses five common text categories and 33 OCR generative tasks, which is the most comprehensive coverage to date. It includes 1,060 samples that consist of high-quality instruction-image-GT pairs, featuring high text density in images, varied generated text amount, diverse image aspect ratios, and bilingual content. To enable holistic evaluation, we introduce OCRGenScore, a unified metric that comprehensively measures model performance across text accuracy, structural consistency, and instruction following on OCRGenBench.

We conduct extensive experiments on OCRGenBench with 19 representative generative models, including both unified and specialized architectures from open-source and closed-source domains. Through quantitative and qualitative evaluation, we distill eight critical findings revealing limitations of existing models and pinpointing potential research directions for visual text synthesis. OCRGenBench fills a critical lack of evaluation benchmarks in the OCR generative field, establishing a new test ground for comprehensively assessing generative models' visual text synthesis capabilities, thereby advancing more accurate, versatile, and generalizable OCR generative systems.

\bibliographystyle{abbrvnat}
\bibliography{reference}

\newpage

\end{document}